\documentclass[a4paper,fleqn]{cas-dc}

\usepackage[authoryear]{natbib}
\usepackage{longtable}
\usepackage{algorithm}
\usepackage{algorithmic}
\usepackage{amsmath,amsfonts,amsthm,amssymb}
\usepackage{hyperref}
\usepackage{multirow}
\usepackage{booktabs}
\usepackage{float}
\usepackage{bm}
\newcommand{\tabincell}[2]{\begin{tabular}{@{}#1@{}}#2\end{tabular}}

\usepackage{threeparttable}
\usepackage{tabularx}
\usepackage{diagbox}
\usepackage{xcolor}
\usepackage{subfigure}
\usepackage{array}
\usepackage{url}
\usepackage{verbatim}
\usepackage{graphicx}

\usepackage{soul, color, xcolor}

\soulregister{\cite}7 
\soulregister{\citep}7 
\soulregister{\citet}7 
\soulregister{\ref}7 
\soulregister{\pageref}7 
\usepackage{wrapfig}
\usepackage{hyperref}

\def\tsc#1{\csdef{#1}{\textsc{\lowercase{#1}}\xspace}}
\tsc{WGM}
\tsc{QE}

\begin{document}
\let\WriteBookmarks\relax
\def\floatpagepagefraction{1}
\def\textpagefraction{.001}

\shorttitle{GELD: A Unified Neural Model for Efficiently Solving Traveling Salesman Problems Across Different Scales}    

\shortauthors{Yubin Xiao et al.}  

\title [mode = title]{GELD: A Unified Neural Model for Efficiently Solving Traveling Salesman Problems Across Different Scales}  



%

\author[1]{Yubin~Xiao}
\author[2]{Di~Wang}
\author[1]{Rui~Cao}
\author[1]{Xuan~Wu}
\cormark[1]
\author[3]{Boyang~Li}
\author[1]{You~Zhou}
\cormark[1]


\affiliation[1]{organization={Key Laboratory of Symbolic Computation and Knowledge Engineering of Ministry of Education, College of Computer Science and Technology, Jilin University},
            city={Changchun},
            postcode={130012},
            country={China}}


\affiliation[2]{organization={Joint NTU-UBC Research Centre of Excellence in Active Living for the Elderly, Nanyang Technological University},
            postcode={639956},
            country={Singapore}}

\affiliation[3]{organization={College of Computing and Data Science, Nanyang Technological University},
            postcode={639956},
            country={Singapore}}

\cortext[1]{Corresponding author}



\begin{abstract}
The Traveling Salesman Problem (TSP) is a well-known combinatorial optimization problem with broad real-world applications. Recent advancements in neural network-based TSP solvers have shown promising results. Nonetheless, these models often struggle to efficiently solve both small- and large-scale TSPs using the same set of pre-trained model parameters, limiting their practical utility. To address this issue, we introduce a novel neural TSP solver named GELD, built upon our proposed broad global assessment and refined local selection framework. Specifically, GELD integrates a lightweight Global-view Encoder (GE) with a heavyweight Local-view Decoder (LD) to enrich embedding representation while accelerating the decision-making process. Moreover, GE incorporates a novel low-complexity attention mechanism, allowing GELD to achieve low inference latency and scalability to larger-scale TSPs. Additionally, we propose a two-stage training strategy that utilizes training instances of different sizes to bolster GELD's generalization ability. Extensive experiments conducted on both synthetic and real-world datasets demonstrate that GELD outperforms seven state-of-the-art models considering both solution quality and inference speed. Furthermore, GELD can be employed as a post-processing method to significantly elevate the quality of the solutions derived by existing neural TSP solvers via spending affordable additional computing time. Notably, GELD is shown as capable of solving TSPs with up to 744,710 nodes, first-of-its-kind to solve this large size TSP without relying on divide-and-conquer strategies to the best of our knowledge. 
\end{abstract}





\maketitle

\section{Introduction}
The Traveling Salesman Problem~(TSP) is one of the most well-known Combinatorial Optimization Problems (COPs) and has extensive real-world applications \citep{Ha2018,Arnold2021}. Due to the practical significance of TSP, many exact, approximate, and heuristic algorithms have been developed over the years. Recently, advances in deep learning have led researchers to develop Neural Networks (NNs) as a kind of viable solvers for TSPs \citep{Yan2022,Wu2024}. Although theoretical guarantees for such networks remain elusive, they tend to produce near-optimal solutions in practice, offering faster inference speed and better generalization than conventional TSP solvers \citep{Bengio2021}. 

Neural TSP solvers often demonstrate excellent performance when trained and tested on small-scale TSPs (e.g., around 100 nodes) \citep{Kwon2020}. However, existing models generally face the following four key limitations: 1)~Generalizing pre-trained models to TSPs of different sizes often results in substantial performance degradation \citep{Joshi2022}. This limitation poses a major obstacle towards deploying these models because real-world TSPs often involve tasks across different scales; 2)~The quadratic time-space complexity ($\mathcal{O}(n^2)$, where $n$ denotes the number of nodes in the underlying TSP) of the standard attention mechanism commonly used in neural TSP solvers restricts their applicability to large-scale TSPs (e.g., over 5,000 nodes); 3)~Further elevation of solution quality, e.g., via sacrificing computing time, is challenging because the NN used in the model often serves as fixed mapping functions from node features to TSP solutions \citep{Xiao2024A}; and 4)~While models utilizing the Divide-and-Conquer~(D\&C) strategy demonstrate high inference speed in solving large-scale TSPs \citep{Zheng2024}, they often yield suboptimal solutions mainly caused by neglecting inter-dependencies among sub-problems during the division process \citep{Luo2024}. Thus, in this work, we investigate the following research question:

\textit{Can a unified pre-trained model, not based on D\&C, effectively solve both small- and large-scale TSPs in a short time period while further elevating solution quality at the cost of affordable time?}

To answer this research question, we introduce GELD, a novel model that integrates a Global-view Encoder (GE) and a Local-view Decoder (LD) to efficiently solve TSPs. Firstly, GELD is built upon our proposed broad global assessment and refined local selection framework (see Section~\ref{Sec32}). Specifically, GELD employs a lightweight GE to capture the topological information across all nodes in the underlying TSP, paired with a heavyweight LD to autoregressively select the most promising node within a local selection range. This dual-perspective approach enriches the embedding representation by integrating both global and local insights while accelerating the selection process by confining the decision space to a smaller, more relevant subset, thereby improving both efficiency and generalization. Secondly, to reduce model complexity and further accelerate inference, we propose a novel Region-Average Linear Attention (RALA) mechanism within GE which operates with $\mathcal{O}(n)$ time-space complexity. RALA partitions the nodes in the underlying TSP into regions and facilitates efficient global information exchange through regional proxies, allowing GELD to solve TSPs in a short time period and scale effectively to larger instances. Thirdly, to further elevate solution quality, we incorporate our proposed idea of diversifying model inputs (see Section~\ref{Sec33}) into GELD's architectural design, enabling the model to function not only as a TSP solver but also as a powerful post-processing method to efficiently exchange affordable computing time for improved solution quality. Finally, to ensure GELD’s robustness across TSPs of varying sizes, we propose a two-stage training strategy, incorporating instances across different scales. This approach further strengthens GELD’s generalization capability, allowing it to solve TSPs efficiently with the same set of pre-trained parameters.

To evaluate the effectiveness of GELD, we conduct extensive experiments on both synthetic and widely adopted benchmarking real-world datasets. The results demonstrate that GELD outperforms seven state-of-the-art (SOTA) models considering both solution quality and inference speed. Furthermore, as a post-processing method, GELD not only significantly enhances the solution quality of baseline models with insignificant additional computing time, but also effectively solves extremely large TSPs (up to 744,710 nodes) when integrated with conventional heuristics. Our findings strongly suggest that GELD is by far the most SOTA model for solving TSPs across different scales.

The key contributions of this work are as follows.

{\romannumeral 1}) To the best of our knowledge, GELD is the first unified model with a single set of pre-trained parameters that effectively solves TSPs across different scales while efficiently enhancing the solution quality.

{\romannumeral 2}) We propose a novel low-complexity encoder-decoder backbone architecture for GELD, enabling low-latency problem-solving and scalability to larger TSP instances.

{\romannumeral 3}) We propose a two-stage training strategy that utilizes instances across different scales to enhance GELD's generalization ability.

{\romannumeral 4}) We show the effectiveness of GELD both as a standalone TSP solver and as a powerful post-processing method that exchanges affordable time for elevated solution quality by conducting extensive experiments.

\section{Related Work}
This section first reviews the NN-based TSP solvers and then introduces recent endeavors aimed at enhancing model generalization.

\subsection{Neural Network-based TSP Solvers}
NN-based methods have shown promising results in solving TSPs and COPs \citep{Sheng2022,Huang2022a}. They can be broadly classified into the following two categories: 1)~Neural construction methods. These methods produce TSP solutions either autoregressively (majority) \citep{Kool2019, Jin2023} or in a one-shot manner (minority) \citep{Xiao2023, Min2023}. For instance, \citet{Kool2019} proposed a well-known Attention Model (AM) for solving TSPs. Moreover, numerous studies extended AM and achieved better solution quality \citep{Kim2022, Kwon2021, Felix2023, Xiao2024}, with POMO \citep{Kwon2020} being the most representative model. 2)~Neural improvement methods. These methods start with initial solutions and employ pre-trained NNs to guide or assist the optimization of heuristics to iteratively improve the solutions \citep{Li2023, Liu2024}. In this line of research, local search \citep{Hudson2022, Ma2023} and evolutionary computation \citep{Ye2023, Kim2024} algorithm are often utilized. 

Despite progress in both categories, these methods typically operate independently. To the best of our knowledge, there does not exist a unified approach capable of both producing and improving TSP solutions. To fill in this gap, in this paper, we propose a unified model that serves as both a standalone TSP solver and a post-processing method to further elevate solution quality. We believe that unifying neural construction methods (solution generation) and neural improvement methods (solution optimization) is both practically significant and necessary for the following key reasons:

\textbf{1) Enhanced Flexibility for Real-world Applications}: A unified model that supports both solution generation and optimization enables efficient adaptation to diverse scenarios, significantly broadening its application scope. For instance, in food delivery route planning, such a model can rapidly generate high-quality routes during peak hours and optimize the routes during off-peak hours for higher solution quality with slightly increased computational time. This flexibility leads to enhanced delivery efficiency and reduced overall costs.

\textbf{2) Cost and Resource Efficiency}: Maintaining a single model with a unified set of parameters for both types of tasks reduces resource requirement, training/optimization time, and deployment complexity compared to separate models. This efficiency is particularly valuable in industries with dynamic demands, such as autonomous mining, smart logistic hubs, automated ports and terminals, etc., enabling faster adaptation and streamlined operations.

\subsection{Generalization of Neural TSP Solvers}
Early studies on neural TSP solver primarily focused on small-scale instances, which limited their applicability to practical and larger-scale scenarios. Recent efforts have sought to extend pre-trained models to larger-scale TSPs, often employing Divide-and-Conquer (D\&C) strategies \citep{Fu2021, Li2021a, Cheng2023,Hou2023, Pan2023, Ye2024, Yu2024, Goh2024}. These models decompose a large-scale problem into multiple smaller sub-problems, solve them individually or in parallel, and then combine the solutions of these sub-problems to construct the complete solution for the original problem. While effective for large-scale TSPs, D\&C-based methods often yield suboptimal solutions because they generally overlook correlations between sub-problems during the  division process \citep{Luo2024}.

Beyond D\&C-based models, alternative learning paradigms, such as diffusion models \citep{Sun2023}, have shown excellent performance in solving large-scale TSPs. Among these non-D\&C-based models, models such as BQ \citep{Drakulic2023} and LEHD \citep{Luo2023} yielded notable results in solving medium-scale TSPs (e.g., with 1,000 nodes) by leveraging the recursion nature of COPs. However, these models struggle to solve TSPs exceeding 1,000 nodes and require significant computing time (see Table~\ref{tab_syn}), limiting their real-world applicability.

To enhance the practicality of neural TSP solvers, we aim to effectively solve both small- and large-scale TSPs using a unified model. This approach is necessary because real-world TSPs may exhibit varying sizes and require well-generalized solvers.

\section{Preliminaries}
In this section, we first detail the TSP setting and the autoregressive mechanisms used in neural TSP solvers. Next, we identify potential generalization issues in neural TSP solvers and outline the motivation behind the framework design of GELD. Finally, we review existing operations that exchange computing time for elevated solution quality and discuss the rationale for diversifying inputs. 

\subsection{TSP Setting and Autoregressive Models}
Our research focuses on the most fundamental Euclidean TSP due to its importance and prevalence in various application domains \citep{Applegate2007, Qiu2022}. We denote a TSP-$n$ instance as a graph with $n$ nodes in the node set $V$, where node $x_i\in\mathbb{R}^{n\times d}$ denotes the $d$-dimensional node coordinates. We define a TSP tour as a permutation of $n$ nodes denoted by $\pi = \{\pi_1,\pi_2,...,\pi_n\}$, where $\pi_i\neq \pi_j, \forall i\neq j$. The goal is to find a feasible solution~$\pi^*$ that minimizes the overall length~$\operatorname{L(\pi^*)}$. 

Autoregressive models are commonly employed to solve TSPs following the Markov Decision Process (MDP). Specifically, at each step $t$, the model whose parameters are collectively denoted as $\theta$, takes an action $a_t$ based on the previously taken actions $a_{1:t-1}$ to choose an unvisited node, until the tour is completed. Given a TSP instance $s$, this process can be factorized into a chain of conditional probabilities as follows:
\begin{equation}\label{eq2}
p_\theta(\pi|s)=\prod\nolimits_{t=1}^{n}p_\theta(a_t|a_{1:t-1},s).
\end{equation}

\subsection{Generalization Issues}\label{Sec32}
Effectively generalizing across TSPs across different scales is a crucial capability for NN-based models \citep{Joshi2022, Zong2022a}. This task is challenging due to the explosive growth in the feasible solution space ($\mathcal{O}(n!)$) as $n$ increases. In autoregressive neural TSP solvers, larger-size instances lead to both increased MDP steps and an expanded decision space (i.e., available nodes) at each step (see~(\ref{eq2})). To better deal with these issues, we propose to confine the decision space at each step to a limited range. Our strategy has certain resemblance to the recent INViT-3V model \citep{Fang2024}, which utilizes multiple local views to solve large-scale TSPs. While INViT-3V excels in solving large-scale TSPs,  its exclusive focus on local information results in suboptimal performance on smaller-scale ones (see Table~\ref{tab_syn}). Conversely, models such as ELG \citep{Gao2024}, which integrate both global and local views, tend to prioritize local information for decision-making without reducing the decision space. Consequently, these models still face challenges in effectively solving large-scale TSPs (see Table~\ref{Tab_real}).

Unlike previous approaches, we introduce a novel \textit{broad global assessment and refined local selection} framework in this paper, which draws inspiration from common decision-making processes in daily life: We often survey adequate relevant information broadly before carefully selecting the most promising option from a few sensible candidates. When applied to solve COPs, this framework involves an initial rough assessment of the entire problem, followed by a zoomed-in focus on the promising candidates, and selection of the most promising one as the action at each decision step. Building upon this idea, we aim to generalize our model to effectively solve TSPs across different scales.

\subsection{Exchange Time for Solution Quality}\label{Sec33}
Neural TSP solvers often utilize a greedy strategy, selecting the node with the highest probability at each MDP step. While computationally efficient, this approach often results in suboptimal solutions \citep{Hottung2022}. To improve solution quality, researchers have proposed various methods, often at the expense of increased computing time. These methods can be broadly categorized into the following two types: 1)~Producing multiple candidate solutions utilizing techniques such as data augmentation \citep{Geisler2022}, multiple rollouts \citep{Kwon2020, Hottung2024}, and various search methods \citep{Choo2022, Garmendia2024}; and 2)~Employing post-processing techniques, such as 2-opt \citep{Sun2023}, monte carlo tree search \citep{Xia2024}, and Re-Construction~(RC) \citep{Luo2023, Luo2024, Ye2024} to improve the quality of initial solutions. Given the versatility and efficiency of these approaches, we primarily employ Beam Search (BS) and RC to balance computing time and solution quality.

BS is a breadth-first search method with a width of $B$ \citep{Kool2019}. It begins with the starting node and incrementally expands the tour by evaluating $B$ potential successors. At each step, BS retains the top-$B$ sub-tours based on their cumulative logarithmic probabilities. 

After obtaining initial solutions, RC randomly selects sub-solutions, reintegrates their node features into the model, and generates new sub-solutions using a greedy strategy. If these new sub-solutions are of higher quality, they replace the current ones. Importantly, RC is fundamentally distinct from the D\&C strategy which decomposes a large problem into multiple smaller sub-problems---a process that may lead to suboptimal solutions. Instead, RC exploits the property that the optimal solution of COPs comprises optimal sub-solutions and improves the overall solution quality by enhancing the quality of these sub-solutions. Thus, RC does not require problem decomposition and operates directly on the initial solution, making it applicable to a wider range of scenarios. Furthermore, when multiple sub-solutions are processed in parallel, referred to as Parallel RC (PRC) \citep{Luo2024}, this parallel approach yields promising results in effectively exchanging computing time for further elevated solution quality. 

We attribute the effectiveness of RC to \textit{the diversification of model inputs}. The rationale behind this is as follows: RC improves solution quality by generating different sub-solutions, which essentially expands the search space. However, NNs are often treated as fixed mapping functions from inputs to outputs. If the model's inputs remain relatively unchanged, the search space is restricted, leading to relatively fixed outputs and limited solution quality improvement possibilities. Therefore, we deem that increasing the diversification of model inputs may enhance the effectiveness of RC. Based on this rationale, we impose the need for diversified inputs during the RC process in our model architectural design. We present the detailed RC process used in our model in Section~\ref{De_rc}.

\begin{figure*}[!t]
\centering
\includegraphics[width=2\columnwidth]{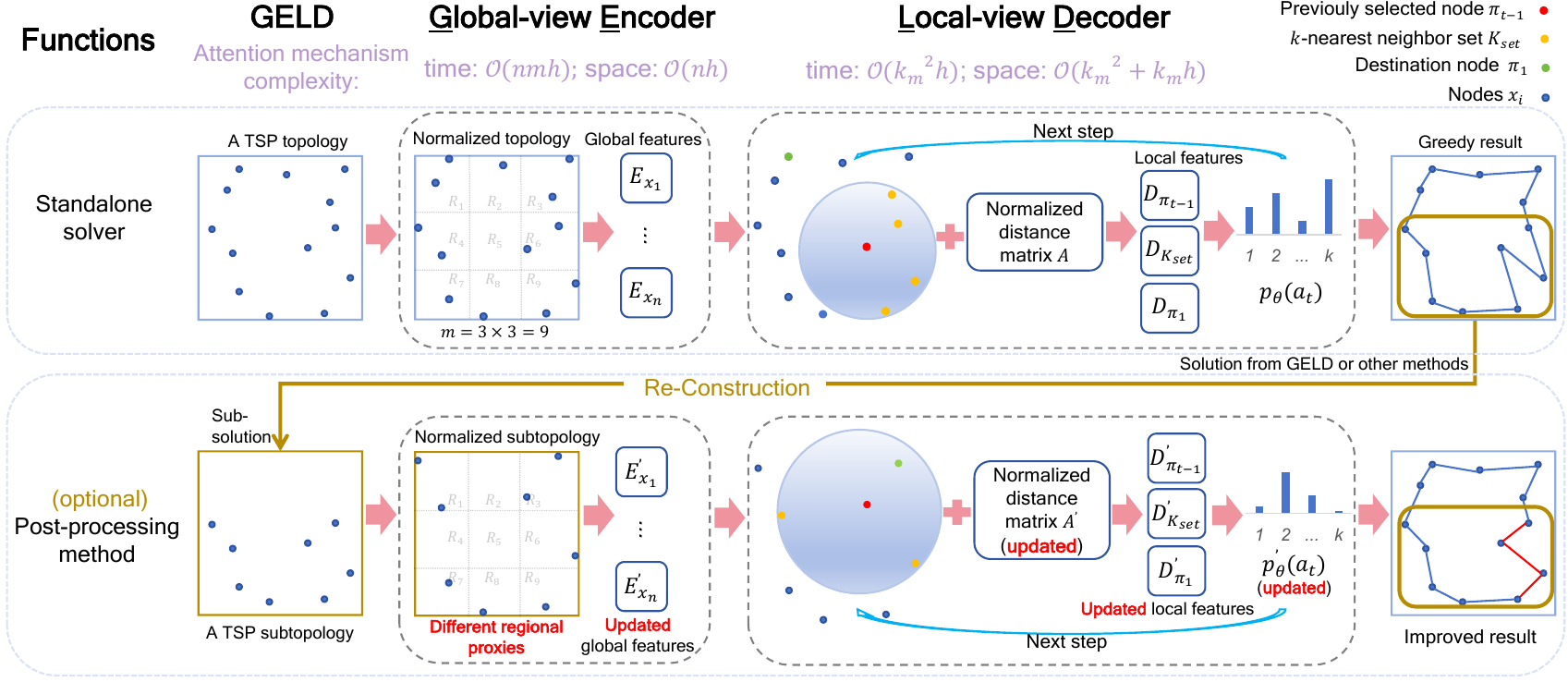}
\caption{Framework of our proposed GELD, which incorporates a low complexity architecture with a GE and an LD. Furthermore, the effectiveness of RC is improved by considering the need for diversified inputs in the model architectural design.}\label{fig_main}
\end{figure*}

\section{\underline{G}lobal-view \underline{E}ncoder and \underline{L}ocal-view \underline{D}ecoder (GELD)}
This section introduces a novel neural TSP solver named GELD. We detail the model architecture, training strategy of GELD, and the used RC process in the following subsections.

\subsection{Architecture of GELD}\label{Arc}
In alignment with the broad global assessment and refined local selection framework, we adopt an encoder-decoder architecture. The encoder captures the topological information across all nodes in the underlying TSP with a global view (\textit{global assessment}), while the decoder employs a local perspective to autoregressively generate the probability distribution for selecting the next node at each step of the MDP (\textit{local selection}). We present the overall framework of GELD in Figure~\ref{fig_main}.

\noindent\textbf{Global-view Encoder}.
To capture global information in the TSP, we account for several distribution patterns, such as the clustered distribution \citep{Jakob2019}, which may only occupy a subset of the graph. Before identifying the patterns, we first normalize the node coordinates $x$ as follows:
\begin{equation}\label{eq3}
    \phi(x) = \frac{x-\min_{x_i\in V}(x_i)}{\max_{x_i, x_j\in V}(x_i-x_j)}.
\end{equation}

Furthermore, during the RC process, the normalization operation alters the node coordinates according to the node changes in node set $V$, which consists of (different) nodes derived from randomly selected sub-solutions, thereby modifying the model input and enhancing the efficacy of RC. Then, we linearly project the normalized coordinates into an $h$-dimensional embedding as follows:
\begin{equation}\label{eq4}
E = \phi(x)W + b, E \in \mathbb{R}^{n\times h},
\end{equation}
where $W$ and $b$ denote the learnable parameters of weights and biases, respectively.

In alignment with the \textit{broad global assessment} aspect of the proposed framework, which involves broadly surveying the relevant TSP information, we utilize \textit{a single}~(\textit{broad}) attention layer to extract \textit{global} features of nodes. Notably, extracting these global features presents challenges because it requires meeting the following three criteria: 1)~Comprehensive coverage of all node information to enable interaction among nodes and facilitate global information transfer; 2)~Low computational complexity to ensure scalability to larger-scale TSPs; and 3)~The ability to obtain global information in a vague manner, allowing for effective diversifying model inputs during the RC process.

Existing models often adopt the standard attention mechanism \citep{Vaswani2017} to facilitate global information transfer, which aids in mapping a query $Q=EW$ to an output using a set of key-value pairs $K=EW$ and $V=EW, Q, K, V\in\mathbb{R}^{n\times h}$ as follows:
\begin{equation}\label{eq5}
    E=\operatorname{\text{Softmax}}(QK^T)V, E \in \mathbb{R}^{n\times h}.
\end{equation}

While the standard attention mechanism delivers strong performance, its quadratic complexity, specifically the time complexity of $\mathcal{O}(n^2h)$ and the space complexity of $\mathcal{O}(n^2+nh)$, limits a model's scalability to larger-scale instances. 

\begin{figure}[!t]
\centering
\includegraphics[width=\linewidth]{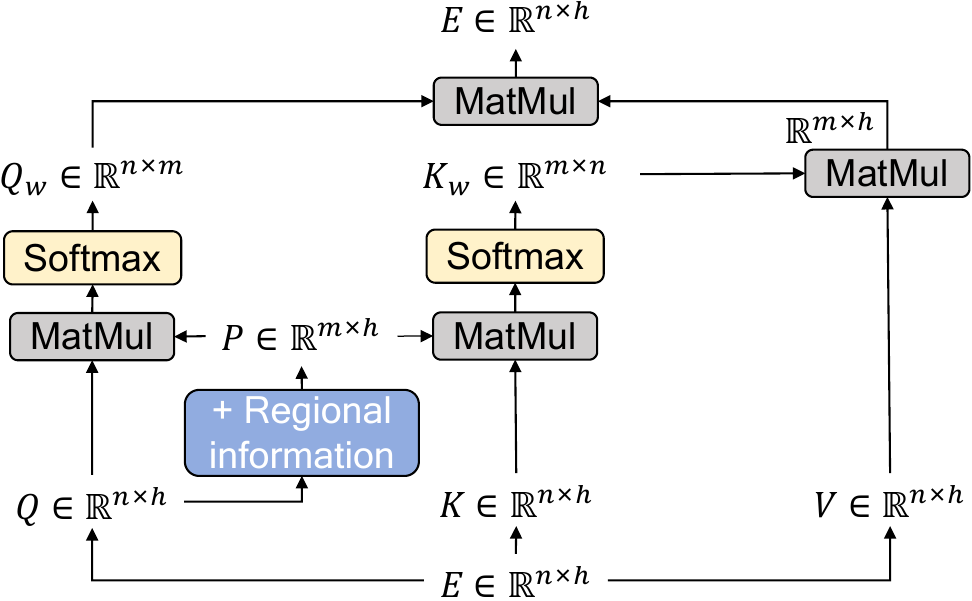}
\caption{Computation process of RALA.}\label{Fig_RALA}
\end{figure}

To meet the aforementioned three criteria, we propose  Region-Average Linear Attention (RALA) that captures global node features with a reduced computational complexity. We present the detailed computation process of RALA in Figure~\ref{Fig_RALA}. Specifically, we first partition all nodes into $m$ regions according to the normalized node coordinates, denoted as $R_1, \dots, R_m$. Here, $m=m_r\cdot m_c$ and $m\ll n, h$, where $m_r, m_c\in\mathbb{Z}^+$ denote the predefined numbers of rows and columns for partitioning, respectively. The derived hyperparameter $m$ controls the granularity of the regional view: a larger value of $m$ may capture more insights of local regions but increases complexity. Specifically, we illustrate this dynamic through two extreme cases as follows:
\begin{itemize}
\item[$\bullet$] \textbf{Case 1: $m$=1.} In this configuration, all nodes in the TSP graph are ground into a single region. This setup minimizes computational complexity, because a single regional representation aggregates the global information by averaging all node embeddings. However, the averaging process inherently smooths out local variations, limiting the model's ability to retain detailed node-specific (local) information.

\item[$\bullet$] \textbf{Case 2: $m$ is large}: When $m$ is sufficiently large, each node is assigned to its own region (while many regions may not contain any node). This configuration maximizes the retention of local information, because each regional representation is directly represented by the individual node embedding. While this setup preserves local details, it significantly increases computational complexity, restricting its applicability to large-scale TSPs.
\end{itemize}
These examples underscore the trade-off inherent in the choice of $m$. Smaller values of $m$ favor computational efficiency and emphasize on global information, while larger values prioritize local details at the expense of scalability. We present the sensitivity analysis of the number of regions $m$ in Section~\ref{Sec_abl}.

Then, we employ regional proxies to facilitate global information exchange among all nodes, thereby meeting the first aforementioned criterion. We compute the embedding of each regional proxy $P_i$ by averaging the query embedding $Q$ of all nodes in this region as follows: 
\begin{equation}\label{eq7}
	 P_i =\begin{cases}
		\frac{1}{n_{R_i}}\sum Q_{x_j}, x_j \in R_i, & \mbox{if } n_{R_i}>0,\\
		0_{1\times h}, & \mbox{otherwise},
	\end{cases}
  i\in \{1, \dots m\},
\end{equation}
where $P \in \mathbb{R}^{m\times h}$, $n_{R_i}$ denotes the number of nodes in region $R_i$, and $Q_{x_i} \in \mathbb{R}^{1\times h}$ denotes the embedding of node $x_i$ in the query $Q$.

Next, we compute the node's query weight score for each region as follows:
\begin{equation}\label{eq8}
Q_w = \operatorname{\text{Softmax}}(QP^T), Q_w\in  \mathbb{R}^{n\times m}.
\end{equation}

Similarly, we compute the regional proxy's key weight score for each node as follows:
\begin{equation}\label{eq9}
K_w = \operatorname{\text{Softmax}}(PK^T), K_w\in  \mathbb{R}^{m\times n}.
\end{equation}

Finally, we update the node features to facilitate the global information transfer as follows:
\begin{equation}\label{eq10}
E = Q_w(K_wV), E \in  \mathbb{R}^{n\times h}.
\end{equation}

Unlike the quadratic complexity of the standard attention mechanism, our proposed RALA achieves a time and space complexity of $\mathcal{O}(nmh)$ and $\mathcal{O}(nh)$, respectively, without introducing any additional learnable parameters. This efficiency makes RALA meet the second aforementioned criterion, capable of solving large-scale instances efficiently. Furthermore, during the RC process, the introduction of normalization operations (see (\ref{eq3})) leads to nodes being assigned to different regions for RALA execution, as illustrated in Figure~\ref{fig_main}. The diversification in regional proxies updates the global features and then enhances the effectiveness of RC, meeting the third aforementioned criterion.

\noindent\textbf{Local-view Decoder}. In alignment with the \textit{refined local selection} aspect of the proposed framework, which selects the most promising option from several candidates, we utilize \textit{multiple}~(\textit{refined}) attention layers within the local-view decoder. Following the decoder design adopted in LEHD \citep{Luo2023} and BQ \citep{Drakulic2023}, we select the most promising node $\pi_{t}$ from a candidate set based on the information from the previously selected node $\pi_{t-1}$ and the destination node $\pi_1$ at MDP step $t$. Unlike LEHD and BQ that consider all available nodes as candidates, we restrict the candidate set to the available $k$-nearest neighbors $K_\textit{set}$ of node $\pi_{t-1}$ (i.e., \textit{local selection}), where $k=\min\{k_m,n_t\}$, with hyperparameter $k_m$ denoting the maximum local selection range and $n_t$ denoting the number of remaining available nodes at step~$t$. This approach reduces the decision space and accelerates the decision-making process (see Table~\ref{Tab_abla}). Formally, we denote the features of nodes $\pi_{t-1}$ and $\pi_1$ and the candidate set $K_\textit{set}$ as $E_{\pi_{t-1}} \in \mathbb{R}^{1\times h}$, $E_{\pi_1} \in \mathbb{R}^{1\times h}$, and $E_{K_\textit{set}} \in\mathbb{R}^{k\times h}$, respectively. We concatenate these features to form the decoder's input at MDP step $t$ as follows:
\begin{equation}\label{eq11}
 D =(E_{\pi_{t-1}}, E_{K_\textit{set}}, \ldots, E_{\pi_1}), D\in\mathbb{R}^{(k+2)\times h}.
\end{equation}

To capture subtle distinctions between the nodes within the local selection range, we employ the attention mechanism used in \citep{Zhou2024} which integrates the distance matrix $A$ among the decoder input nodes. Notably, during the RC process, the normalization operation alters $A$, further enhancing feature diversity and potentially improving solution quality. Additionally, to mitigate potential value overflows due to repeated exponential operations, we incorporate RMSNorm \citep{Biao2019} into the attention mechanism. The time and space complexity of the attention mechanism in our decoder is $\mathcal{O}(k_m^{\ \ 2}h)$ and $\mathcal{O}(k_m^{\ \ 2}+k_mh)$, respectively.

After refining the local node features through multiple attention layers, we compute the probability distribution of nodes in candidate set $K_\textit{set}$ being selected at MDP step $t$ as follows: 
\begin{equation}\label{eq15}
p_\theta(a_t)=\operatorname{\text{Softmax}}\left(D_{x_i}W\odot \begin{cases}
    1, & \text{if}~x_i \in  K_\textit{set},\\
    -\infty, & \text{ otherwise,} 
\end{cases}\right ), 
\end{equation}
where $p_\theta(a_t)\in\mathbb{R}^{k}$, $D_{x_i}$ denotes the features of node $x_i$, and $\odot$ denotes the element-wise multiplication.

\subsection{Training Strategy of GELD}
Existing neural TSP solvers typically rely on Supervised Learning (SL) \citep{Luo2023, Drakulic2023}, Reinforcement Learning (RL) \citep{Gao2024,Fang2024}, or Self-Improvement Learning (SIL) \citep{Luo2024, Pirnay2024} for model training. We choose not to use RL due to its requirement of generating a complete solution before calibrating the reward, which normally requires a large amount of computational resources. Inspired by recent advancements in fine-tuning large models \citep{Han2024}, we propose a two-stage training approach. The first stage involves SL training on small-scale instances, followed by SIL training on larger instances. For the first stage, we adopt the same SL method used by \citep{Drakulic2023} and \citep{Luo2023} and utilize the publicly available training dataset contributed by \citep{Luo2023} to ensure fair comparisons in all relevant experiments.

However, the experimental results reveal that models (e.g., GD \citep{Pirnay2024}) trained on small-scale TSPs exhibit limited generalization capacity on larger-scale TSPs (see Table~\ref{tab_syn}). We hypothesize that this limitation arises because NN-based models typically map inputs to outputs in a fixed manner. When the node distribution in the test data significantly differs from that in the training data, the model struggles to generalize effectively. Thus, we expand the training data size in the second stage to mitigate the model's reduced effectiveness in solving larger instances. Because obtaining ground-truth labels for large-scale TSPs is costly, we use the pseudo-labels generated by GELD for self-improvement learning in the second stage. We introduce the mechanisms of each training stage as follows.

\noindent\textbf{SL Training on Small-scale TSPs}.
We define TSPs with fewer than $k_m$ (i.e., the maximum local selection range) nodes as small-scale TSPs. For a TSP-$n$ training instance $s$, we employ the cross-entropy function to maximize the probability of selecting the optimal action at each step as follows:
\begin{equation}\label{eq13}
\mathcal{L}(\theta|s)=-\sum\nolimits^{n}_{i=1}y_i\log(p_\theta(i)),
\end{equation}
where $n\leq k_m$, $y_i\in\{0, 1\}$ denotes the ground-truth label, indicating whether node $x_i$ should be selected at the current step, and $p_\theta(i)$ denotes the probability of selecting node $x_i$. 

\begin{algorithm}[t]
	\caption{Two-stage training strategy of GELD.} \label{alg1}
	\begin{algorithmic}[1]
		\REQUIRE small-scale TSP-$k_m$ dataset $\textit{data}_s$, training batch size $n_{\textit{bs}}^t$, maximum training size $n_\textit{max}$, epoch numbers $n_{e1}$ and $n_{e2}$ for the first stage and the second stage, respectively, training termination hyperparameters $t_\textit{max}, \epsilon$, and $t_\textit{imp}$. 
		\STATE Initialize $\theta$ \\ \textcolor{gray}{\#\textit{The first-stage SL training on small-scale TSPs}}
		\FOR {$\textit{epoch}$ in $1,...,n_{e1}$}  
		\STATE $\textit{data}_1,\textit{label}_1 \gets$ TSP-$n$ instances from $\textit{data}_s$, where $n \leq k_m$
		\STATE $\theta \gets$ \textbf{GELD}($\theta$, $\textit{data}_1$, $\textit{label}_1$)
		\ENDFOR \\ \textcolor{gray}{\#\textit{The second-stage SIL training on large-scale TSPs}}
  	\FOR {$\textit{epoch}$ in $1,...,n_{e2}$}
        \STATE $l_\textit{scale} \gets k_m + \textit{epoch} \cdot (n_\textit{max} - k_m) \mid n_{e2}$
        \STATE $\textit{data}_2 \gets$ Randomly generate $n_{\textit{bs}}^t$ TSP-$l_\textit{scale}$ instances
        \STATE $\textit{len}_G, \_ \gets$ $\operatorname{\text{Greedy strategy}}(\textbf{GELD}, \textit{data}_2)$
        \STATE $\textit{len}_I, \textit{solution} \gets$ PRC(BS$(\textbf{GELD}, \textit{data}_2$))
        \STATE $t_1 \gets 0$, $t_2 \gets 0$
        \WHILE{$t_1 < t_\textit{max}$ and $\frac{\textit{len}_G}{\textit{len}_I} - 1 > \epsilon$ and $t_2 < t_\textit{imp}$}
        \STATE $\textit{data}_1, \textit{label}_1 \gets$ Randomly sample $n_{\textit{bs}}^t$ TSP-$k_m$ instances from $\textit{data}_s$
        \STATE $\textit{data}, \textit{label} \gets \{\textit{data}_2, \textit{solution}\} \cup \{\textit{data}_1, \textit{label}_1\}$
        \STATE $\theta \gets$ \textbf{GELD}($\theta$, $\textit{data}$, $\textit{label}$)
        \STATE $\textit{len}_G, \_ \gets$ $\operatorname{\text{Greedy strategy}}(\textbf{GELD}, \textit{data}_2)$
        \STATE $\textit{len}_{I_\textit{tmp}}, \textit{solution}_\textit{tmp} \gets$ PRC(BS$(\textbf{GELD}, \textit{data}_2$))
        \IF{$\textit{len}_{I_\textit{tmp}} < \textit{len}_I$}
        \STATE $t_2 \gets 0$, $\textit{len}_I \gets \textit{len}_{I_\textit{tmp}}$, $\textit{solution} \gets \textit{solution}_\textit{tmp}$
        \ELSE
        \STATE $t_2 \gets t_2 + 1$
        \ENDIF
        \STATE $t_1 \gets t_1 + 1$
        \ENDWHILE
		\ENDFOR
	\end{algorithmic}
\end{algorithm}

\noindent\textbf{SIL Training on Large-scale TSPs}. After the first training stage, the model exhibits preliminary generalization capability for solving large-scale TSPs. In this second stage, we enhance the model's generalization ability by applying SIL using larger instances, adhering to a curriculum learning strategy that progressively scales the training instances from the small-scale size $k_m$ to a predefined maximum training size $n_\textit{max}$. Specifically, in each training epoch, we randomly generate a batch of $n_\textit{bs}^t$ training instances and apply both BS and PRC to obtain improved solutions (over those produced by the greedy strategy) as pseudo-labels for training. The epoch concludes when any of the following three conditions is met: 1)~The maximum number $t_\textit{max}$ of training iterations per batch is reached; 2)~The gap between the greedy and improved solutions falls below a predefined threshold $\epsilon$; or 3)~There is no further improvement in solution quality after $t_\textit{imp}$ iterations. Furthermore, to prevent overfitting to large-scale problems, we incorporate $n_\textit{bs}^t$ labeled small-scale TSP-$k_m$ instances into the training set at each epoch.

Given a solution $\pi = \{\pi_1,\pi_2,...,\pi_n\}$, either ground-truth labels adopted in SL training or pseudo labels adopted in SIL training, we randomly select a partial solution for model training, e.g., $ \{\pi_i,\pi_{i+1},...,\pi_{i+j}\}$, where $j > 2$. Furthermore, we present the overall two-stage training strategy of GELD in Algorithm~\ref{alg1}. 

\subsection{Detailed Re-Construction Process} \label{De_rc}
This subsection presents the detailed RC process employed in our study, comprising two main steps. Firstly, after obtaining the initial solutions, denoted as $\pi = \{\pi_1,\pi_2,...,\pi_n\}$, RC randomly selects a starting index $i$ and a sub-solution length $j$ to form a sub-solution $\{\pi_i,\pi_{i+1},...,\pi_{i+j}\}$, where $i\in\{1, \dots, n\}$ and $j > 2$. This condition ensures that the sub-solution length is sufficient to impact the outcome,  as sub-problems smaller than size 4 do not alter the sub-solutions during the RC process (i.e., there must be at least two nodes in the candidate set $K_\textit{set}$ for selection). Because the TSP solution $\pi$ forms a cyclic sequence, i.e., $\pi_{n+i}=\pi_i$, RC adapts the sampling direction based on the iteration count, alternating between clockwise (${\pi_i, \pi_{i+1}, \dots, \pi_{i+j}}$) and counterclockwise (${\pi_i, \pi_{i-1}, \dots, \pi_{i-j}}$). To further enhance model input diversity, the solution sequence is shifted by a randomly selected offset $n_\epsilon$ $\in\{1,\dots, n\}$. In the subsequent step, RC reintegrates the selected node features into the model. To introduce additional model input diversity, we randomly apply one of the $\times$8 data augmentation techniques proposed in \citep{Kwon2020}, such as rotating the TSP topology by 90 degrees. The model then generates new sub-solutions using a greedy strategy. If these newly generated sub-solutions outperform the existing ones, they replace the current sub-solutions.

In this study, we leverage GELD's ability to efficiently solve TSPs across different scales. After obtaining an initial solution, we input sub-problems into GELD based on randomly selected sub-solutions to improve them, thereby establishing an effective post-processing method to enhance solution quality without significantly increasing computation time. The source code of GELD is available online\footnote{URL: \href{https://github.com/xybFight/GELD}{https://github.com/xybFight/GELD}.}.

\section{Experimental Results}
This section presents a set of comprehensive evaluations of our proposed GELD. We start by providing an overview of experimental setups. Then, we assess the performance of GELD as a standalone TSP solver and as a post-processing method. Finally, we conduct ablation studies on the proposed architecture and training strategy of GELD.

\subsection{Experimental setups}
In this subsection, we provide a detailed description of the hyperparameter configuration, datasets used, baseline methods, and evaluation metrics.

\subsubsection{Hyperparameter Configuration}\label{sec_HC}
We follow the convention and focus on the 2-dimensional TSP ($d=2$) \citep{Kwon2020}. Our proposed GELD comprises $1$ (\textit{broad}) global-view encoder layer and $6$ (\textit{refined}) local-view decoder layers, each with a hidden dimension of $h=128$ and $8$ attention heads, following \citep{Luo2023}. To balance performance and computational complexity, we set the numbers of rows and columns to $m_r= m_c=3$, resulting in $m=m_r\cdot m_c=3\times3=9$ regions. For fair comparisons with SL-trained models \citep{Drakulic2023, Luo2023}, we adhere to the same training scale, with the small-scale size set to $k_m=100$. While increasing the maximum training size $n_\textit{max}$ intuitively improves model generalization, it also increases computational costs. To strike a balance, we set $n_\textit{max}$ to 1,000. In the first training stage, we utilize the publicly available training dataset $\textit{data}_s$ from \citep{Luo2023} for fair comparisons. The learning rate is set to 1e--4 with a decay rate of 0.97. In the second training stage, the training termination hyperparameters are set to $t_\textit{max}=5, \epsilon=$1e--3, and $t_\textit{imp}=3$. The learning rate is adjusted to 1e--5, and the training batch size $n_{\textit{bs}}^t$ is set to $64$. The width of BS and iteration number of PRC are set to $16$ and 1,000, respectively. All training instances are randomly generated by sampling the node locations based on the uniform distribution pattern. To control the overall training time---approximately 20 hours for the first stage and 31 hours for the second stage---we set the number of epochs $n_{e1}=n_{e2}=50$ for both the first and second stages. All experiments are conducted on a computer equipped with an Intel(R) Core(TM) i9-12900K CPU and an NVIDIA RTX 4090 GPU (24GB).

\subsubsection{Datasets} \label{sec_DC}
We conduct a comprehensive evaluation of model performance using both synthetic datasets and widely recognized real-world benchmark datasets.

\noindent\textbf{Synthetic Datasets.} For the synthetic data, we generate TSP instances across different scales and distributions. Specifically, we synthesize $20$ subsets of TSP instances, encompassing four distribution patterns (namely uniform, clustered, explosion, and implosion) across five scales (namely 100, 500, 1,000, 5,000, and 10,000 nodes), following \citep{Fang2024, Jakob2019}. The number of instances per subset is determined by the scale, comprising 200 instances for TSP-100, TSP-500, and TSP-1000, and 20 instances for TSP-5000 and TSP-10000.

\noindent\textbf{Real-world Datasets.} To assess the model's performance in real-world scenarios, we utilize the widely recognized TSPLIB and World TSP datasets as benchmarks. For TSPLIB, we include all symmetric instances from TSPLIB95\footnote{URL: \href{http://comopt.ifi.uni-heidelberg.de/software/TSPLIB95/}{http://comopt.ifi.uni-heidelberg.de/software/TSPLIB95/}} with nodes represented as Euclidean 2D coordinates, covering 77 instances with sizes ranging from 51 to 18,512 nodes. For World TSP, we include all symmetric instances from National TSPs\footnote{URL: \href{https://www.math.uwaterloo.ca/tsp/world/countries.html}{https://www.math.uwaterloo.ca/tsp/world/countries.html}}, also represented as Euclidean 2D coordinates, comprising 27 instances with sizes ranging from 29 to 71,009 nodes.

\noindent\textbf{Extremely Large-scale TSP Instances.}
To assess the performance of GELD as a post-processing method in extremely large-scale scenarios, we utilize the four largest TSP instances from the VLSI dataset\footnote{URL: \href{https://www.math.uwaterloo.ca/tsp/vlsi/page11.html}{https://www.math.uwaterloo.ca/tsp/vlsi/page11.html}} within the World TSP collection, which include TSP instances with sizes ranging from 104,815 to 744,710 nodes.

\subsubsection{Baseline Methods} \label{sec_BM}
Considering the computational efficiency, we categorize the baseline methods into two groups. Firstly, we compare GELD with conventional algorithms and three SOTA models based on heatmap, using synthetic TSP instances with the uniform distribution. The conventional algorithm include LKH3 \citep{Helsgaun2017}, Concorde \citep{Applegate2007}, Variable Neighborhood Search (VNS) \citep{MLADENOVIC19971097}, Ant Colony Optimization (ACO) \citep{BLUM2005353}, Random Insertion~(RI)~\citep{Azar_1994}, and the Nearest Neighbor+2-opt (NN+2-opt) method. The heatmap-based models include DIFUSCO \citep{Sun2023}, T2T \citep{Li2023}, and DEITSP \citep{Wang2025}. To retain a reasonable computing time consumption, we solve the TSP-{100, 500, 1000} instances using LKH3 with 20,000 iterations over 10 runs, while solving the TSP-{5000, 10000} instances with 20,000 iterations in a single run. For Concorde and Gurobi, the default settings are used. For VNS, we set the maximum number of neighborhood searches to 10. For ACO, we configure a population size of 20 and set the number of iterations to 1,000. For NN+2-opt, the nearest neighbor algorithm generates the initial solution, followed by 2-opt optimization with 1,000 iterations. For the heatmap-based models, we utilize their publicly available pre-trained parameters (trained on 100-node instances) and adopt the default inference settings with 1,000 iterations for the 2-opt optimization.

Secondly, to evaluate generalization performance of a pre-trained model across both small- and large-scale TSPs, we select seven baseline models that have demonstrated SOTA performance across various scales with a single checkpoint. These models include 1)~RL-based models: Omni-TSP \citep{Zhou2023}, ELG \citep{Gao2024}, INViT-3V \citep{Fang2024}, and UDC \citep{Zheng2024}; 2)~SL-based models: LEHD \citep{Luo2023} and BQ \citep{Drakulic2023}; and 3)~SIL-based model: GD \citep{Pirnay2024}. We follow these model's default inference strategy (Omni-TSP, ELG, and UDC adopt greedy multiple rollouts \citep{Kwon2020}, INViT-3V adopts greedy multiple rollouts with data augment technique, while others adopt  the greedy strategy). All baseline models are trained on a uniform distribution pattern, except for Omni-TSP, which was trained on diverse distribution patterns due to the adopted mixed training strategy. Among these, UDC utilizes a D\&C strategy, whereas the others are non-D\&C neural TSP solvers. For the comparative experiments, we adopt the publicly available pre-trained parameters and default settings for all models, with two exceptions: For INViT-3V, we adjust the configuration to handle multiple instances simultaneously, rather than the originally designed single-instance setup, to reduce execution time and ensure a fair comparison; For UDC, we set the hyperparameter values to $x$=250 and $\alpha$=1 in all relevant experiments for a fair comparison. Furthermore, for a fair comparison in terms of computational efficiency, we report results only for the two baseline models---LEHD and BQ---combined with the greedy search strategy.
\begin{table}[!t]
\centering
\caption{Solution length computed by LKH3 on the synthetic dataset}
\label{tab:LKH}
\begin{tabular}{lcccc}
\toprule
& uniform & clustered & explosion & implosion\\
\midrule
TSP-100 &  7.8693 &  5.3876 & 6.5397 & 7.1135 \\
TSP-500 &  16.5601 & 10.3447 & 12.0101 & 14.4128 \\
TSP-1000 & 23.2215 & 14.0982 & 16.0543 & 20.1932 \\
TSP-5000 & 50.9830 & 28.8359 & 31.9792 & 45.0435 \\
TSP-10000 & 73.1436 & 40.2628 & 41.2801 & 63.7273 \\
\bottomrule
\end{tabular}
\end{table}

\subsubsection{Evaluation Metrics} \label{sec_EM}
For all baselines and GELD, we report the average gap to the optimal or the best performing near-optimal solutions. The solutions for synthetic datasets are computed using LKH3 \citep{Helsgaun2017}, while for real-world datasets, we use the best known solutions.  We present the solution length computed by LKH on the synthetic dataset in Table~\ref{tab:LKH}. The gap for each TSP instance is computed as follows:
\begin{equation}
    \text{gap}=\frac{\operatorname{L}(\pi^{\textit{model}})-\operatorname{L}(\pi^{\textit{opt}})}{\operatorname{L}(\pi^{\textit{opt}})}\times100\%,
\end{equation}
where $\pi^{\textit{model}}$ denotes the solution produced by the model and $\pi^{\textit{opt}}$ denotes the (near-)optimal solution. Furthermore, we report the inference time for each baseline method across all datasets. To ensure a fair comparison of inference time for synthetic datasets, we intend to maintain an equal batch size for all models. However, due to the GPU memory constraint (24GB), we use the maximum batch size $n_{\textit{bs}}$ that each model can solve simultaneously. This batch size, reflecting the model's parallel processing capability, serves as a practical measure of inference efficiency under real-world, resource-constrained conditions.

\begin{table*}[!t]
  \centering
    \caption{Performance comparisons with conventional algorithms and heatmap-based models (with 2-opt optimization) on synthetic TSP instances of the uniform distribution} \label{Tab_more}
    \resizebox{2.1\columnwidth}{!}{
    \begin{tabular}{l|cc|cc|cc|cc|cc|c}
    \toprule
    \multicolumn{1}{c|}{\multirow{2}{*}{Method}} & \multicolumn{2}{c|}{\textbf{TSP-100 (200)}} & \multicolumn{2}{c|}{\textbf{TSP-500 (200)}} & \multicolumn{2}{c|}{\textbf{TSP-1000 (200)}} & \multicolumn{2}{c|}{\textbf{TSP-5000 (20)}} & \multicolumn{2}{c|}{\textbf{TSP-10000 (20)}} & \multirow{2}{*}{\tabincell{l}{\textbf{Average}\\ \textbf{gap(\%)}$\downarrow$}
     }\\
    
    & gap(\%)$\downarrow$ & time$\downarrow$, $n_{\textit{bs}}\uparrow$ & gap(\%)$\downarrow$ & time$\downarrow$, $n_{\textit{bs}}\uparrow$ & gap(\%)$\downarrow$ & time$\downarrow$, $n_{\textit{bs}}\uparrow$ & gap(\%)$\downarrow$ & time$\downarrow$, $n_{\textit{bs}}\uparrow$ & gap(\%)$\downarrow$ & time$\downarrow$, $n_{\textit{bs}}\uparrow$ \\
    \cmidrule{1-12}
    LKH3 & 0.00 & 2.7m, 1 & 0.00 & 3.7h, 1 & 0.00 & 15.2h, 1& 0.00 & 1.7h, 1 & 0.00 & 1.3d, 1 & 0.00\\
    Concorde & 0.00 & 8.9m, 1 & -0.04 & 52.0m, 1 & -0.29 & 8.5h, 1& \multicolumn{2}{c|}{N.A.} & \multicolumn{2}{c|}{N.A.} & -\\
    RI & 9.72 & 7.4s, 1 & 12.43 & 35.2s, 1 & 12.66 & 1.7m, 1 & 13.80 & 3.9m & 11.92 & 15.8m & 12.12\\
    VNS & 2.27 & 19.7h, 1 & \multicolumn{2}{c|}{N.A.} & \multicolumn{2}{c|}{N.A.} & \multicolumn{2}{c|}{N.A.} & \multicolumn{2}{c|}{N.A.}&- \\
    ACO & 8.86 & 1.0d, 1& \multicolumn{2}{c|}{N.A.} & \multicolumn{2}{c|}{N.A.} & \multicolumn{2}{c|}{N.A.} & \multicolumn{2}{c|}{N.A.}&- \\
    NN+2-opt & 5.69 & 3.7s, 1& 5.55 & 9.4s, 1 & 5.24 & 34.9s, 1 & 5.53 & 2.7m, 1 & 4.32 & 13.5m, 1 & 5.27\\
    \hline
    DIFUSCO+G (NeurIPS'23)  & 0.28 & 33.6s, 1& 3.12 & 3.3m, 1 & 3.55 & 11.2m, 1 & 5.86 & 8.9m, 1 & 5.66 & 26.9m, 1 & 3.70\\
    DIFUSCO+S (NeurIPS'23) & 0.02 & 1.7m, 16& 1.93 & 10.8m, 16 & 2.66 & 41.9m, 16 & 7.72 & 1.2h, 16 & 14.44 & 4.2h, 16 & 5.35\\
    T2T+G (NeurIPS'23) & 0.12& 4.0m, 1& 2.30 & 6.8m, 1 & 3.33 & 15.5m, 1 &  \multicolumn{2}{c|}{OOM} & \multicolumn{2}{c|}{OOM} & -\\
    T2T+S (NeurIPS'23) & 0.04& 7.2m, 4& 1.91 & 25.4m, 4 & 2.72 & 53.63m, 4 &  \multicolumn{2}{c|}{OOM} & \multicolumn{2}{c|}{OOM} & -\\
    DEITSP+G (KDD'25) & 0.69 & 2.3s, 1 & 3.20 & 9.1m, 1 & 4.13 & 3.4m, 1 & \multicolumn{2}{c|}{OOM} & \multicolumn{2}{c|}{OOM} & - \\
    DEITSP+M (KDD'25) & 0.16 & 18.6s, 1 & 2.01 & 1.4h, 1 & 3.37 & 43.2m, 1 & \multicolumn{2}{c|}{OOM} & \multicolumn{2}{c|}{OOM} & - \\
    \hline
   GELD+G (Ours) & 1.11 & 0.6s, 200& 2.39 & 1.8s, 200 & 2.94 & 3.6s, 200 & 7.62 & 10.8s, 20 & 9.33 & 21.6s, 20 & 4.68\\
    GELD+S$^*$ (Ours) & \textbf{0.06} & 19.2s, 200& \textbf{0.52} & 1.6m, 200 & \textbf{0.58} & 3.7m, 200 & \textbf{2.77} & 1.8m, 20 & \textbf{2.38} & 3.9m, 20 & \textbf{1.26}\\
     \bottomrule
     \end{tabular}}
         \begin{tablenotes}
    \item[1] Symbol ``G", ``S", ``M", and ``S$^*$" 
    denote the greedy strategy, sampling operation used in \citep{Sun2023}, 16 iterations used in \citep{Wang2025}, and the combination of BS with a width of 16 and PRC with 1,000 iterations, respectively. Symbol “OOM” (Out of Memory) is used to indicate cases where the model fails to solve TSPs in the set due to the GPU memory constraint. Symbol “N.A.” is used to indicate that the method exceeds the time limit (e.g., two days).
    \end{tablenotes}
\end{table*}

\subsection{GELD as a Standalone TSP Solver}
Firstly, we compare GELD with conventional algorithms and three SOTA heatmap based models. We present these comparison results on the synthetic TSP instances of the uniform distribution in Table~\ref{Tab_more}. As shown, the performance of heatmap-based models are heavily dependent on the iterative optimization process of 2-opt that (often) specifically tailored to TSP, while our method does not. More importantly, our proposed GELD outperforms these models and conventional algorithms (except for the SOTA LKH3 algorithm) across all problem scales in terms of both solution quality and computational efficiency.

Secondly, we analyze the performance of GELD as a standalone TSP solver on synthetic and real-world datasets, respectively.\\
\noindent\textbf{Synthetic Datasets.} We present the performance comparison of GELD against baselines on synthetic datasets in Table~\ref{tab_syn}. The results indicate that all models, including ours, exhibit performance degradation when generalizing to TSPs across different scales and distribution patterns. This finding highlights the critical need for further research on size generalization. Despite the overall trend of declining performance, our proposed GELD, when paired with the greedy strategy, achieves solution quality on-par with the SOTA INViT-3V model, which employs greedy multiple rollouts and data augment techniques. Moreover, GELD offers a significant advantage in inference speed, consistently outperforming other models across different scales, except for TSP-100. This can be attributed to the efficient, low time complexity backbone architecture of our model (see Figure~\ref{fig_main}). Furthermore, when integrated with BS and PRC, GELD achieves the highest solution quality across all scales and patterns. This superior performance arises from its design, which incorporates diversified model inputs to enhance the effectiveness of RC (see Section~\ref{De_rc}). Additionally, GELD's ability to process all $n_\textit{bs}$ test instances simultaneously across all scales makes it particularly well-suited for practical applications with limited computing resources.

Compared to Omni-TSP, which is trained on diverse distribution patterns, our model, trained on a single (uniform) distribution pattern, shows significantly better cross-distribution generalization performance. However, we do observe a slight performance decrease on instances of non-training distribution patterns. Incorporating cross-distribution instances during training may further enhance our model's performance in this regard, which we plan to explore in future research.

Finally, the INViT-3V model, which relies solely on multiple local views, performs less effectively on small-scale datasets (TSP-\{100, 500\}) compared to large-scale ones (TSP-\{5000, 10000\}). In contrast, our model with a global-view encoder performs well across datasets of all sizes, underscoring the importance of integrating global information to achieve robust performance.\\

\noindent\textbf{Real-world Datasets.}
We present the performance comparison of GELD against baselines on real-world datasets in Table~\ref{Tab_real}. For clarity, the experimental results are grouped by the scale, with detailed performance presented in Appendix~\ref{real_res}. The results demonstrate that GELD consistently outperforms baseline models across all sets of TSP instances in terms of both solution quality and inference speed. Additionally, due to the GPU memory constraint (24GB), all baseline models are unable to solve certain large-scale TSP instances, whereas our model successfully solves all instances. This advantage is attributed to the low space complexity of our model's backbone architecture, again underscoring its suitability for practical applications with limited computing resources.

\begin{table*}[!t]
\small
  \centering
    \caption{Performance comparisons on synthetic TSPs of different sizes and distribution patterns}\label{tab_syn}
    \resizebox{2\columnwidth}{!}{
    \begin{tabular}{l|l|cc|cc|cc|cc|cc|c}
    \toprule
     &  \multirow{2}{*}{\textbf{Model}} & \multicolumn{2}{c|}{\textbf{TSP-100 (200)}} & \multicolumn{2}{c}{\textbf{TSP-500 (200)}} & \multicolumn{2}{c|}{\textbf{TSP-1000 (200)}} & \multicolumn{2}{c|}{\textbf{TSP-5000 (20)}} & \multicolumn{2}{c|}{\textbf{TSP-10000 (20)}} & \multirow{2}{*}{\tabincell{l}{\textbf{Average}\\ \textbf{gap(\%)}$\downarrow$}
     }\\
    & & gap(\%)$\downarrow$ & time$\downarrow$, $n_{\textit{bs}}\uparrow$ & gap(\%)$\downarrow$ & time$\downarrow$, $n_{\textit{bs}}\uparrow$ & gap(\%)$\downarrow$ & time$\downarrow$, $n_{\textit{bs}}\uparrow$ & gap(\%)$\downarrow$ & time$\downarrow$, $n_{\textit{bs}}\uparrow$ & gap(\%)$\downarrow$ & time$\downarrow$, $n_{\textit{bs}}\uparrow$ \\
    \cmidrule{1-13}
    \multirow{13}{*}{\rotatebox{90}{\textbf{uniform}}}& LKH3 & - & 2.7m, 1 & - & 3.7h, 1 & - & 15.2h, 1 & - & 1.7h, 1 & - & 1.3d, 1 & -\\
    \cmidrule{2-13}
    & Omni-TSP (ICML'23) & 2.22 & \textbf{0.3s}, 200 &7.80 & 9.6s, 200 & 19.56 & 1.2m, 100 & 49.43 & 16.1m, 5 & 61.39 & 2.0h, 1 &28.09 \\
    & LEHD (NeurIPS'23)  & 0.67 & 0.7s, 200 &1.58 & 16.2s, 200 & 2.76 & 1.8m, 100 & 15.80 & 18.2m, 5 & 24.10 & 2.3h, 1 &8.96 \\
    & BQ (NeurIPS'23)  & 5.37 &1.5s, 200 & 3.86 & 1.3m, 200 & 3.82 & 9.3m, 100 & 12.68 & 1.9h, 5 & 18.74 & 13.5h, 1 & 8.85 \\
    & ELG (IJCAI'24) & 0.58 & 0.5s, 200 & 8.80 & 4.2s, 200 & 12.22 & 15.6s, 200 & 18.84 & 40.5s, 5 & 18.32 & 3.7m, 2 & 11.77 \\
    & INViT-3V (ICML'24) & 1.47 & 15.2s, 200 & 4.26 & 1.5m, 200 & 4.96 & 3.1m, 200 & 6.60 & 4.4m, 20 & 4.80 & 6.5m, 20 & 4.42 \\
    & GD (TMLR'24) & 0.72 & 3.1s, 200 & 2.25 & 36.4s, 200 & 4.26 & 3.2m, 200 & 60.26 & 26.7m, 20 & 198.65 & 3.4h, 4 & 53.22 \\
    & UDC$^\S$ (NeurIPS'24) & 0.40 & 8.7s, 200 & 2.15 & 28.5s, 200 & 2.06 & 57.2s, 100 & 6.99 & 29.7s, 20 & 8.73 & 2.4m, 1 & 4.07 \\
    
    \cmidrule{2-13}
    & GELD + G (Ours)  & 1.11 & 0.6s, 200 & 2.39 & \textbf{1.8s}, 200 & 2.94 & \textbf{3.6s}, 200 & 7.62 & \textbf{10.8s}, 20 & 9.33 & \textbf{21.6s}, 20 & 4.68 \\
    & GELD + S$^*$ (Ours)  & \textbf{0.06} & 19.2s, 200 & \textbf{0.52} & 1.6m, 200 & \textbf{0.58} & 3.7m, 200 & \textbf{2.77} & 1.8m, 20 & \textbf{2.38} & 3.9m, 20 & \textbf{1.26} \\
    \cmidrule{1-13}
    \multirow{13}{*}{\rotatebox{90}{\textbf{clustered}}}& LKH3 & - & 3.1m, 1 & - & 4.1h, 1 & - & 16.1h, 1 & - & 3.0h, 1 & - & 1.5d, 1 & -\\
    \cmidrule{2-13}
    & Omni-TSP (ICML'23) & 2.37 & \textbf{0.3s}, 200 & 9.82 & 9.6s, 200 & 21.20 & 1.2m, 100 & 54.49 & 16.1m, 5 & 71.60 & 2.0h, 1 &26.56 \\
    & LEHD (NeurIPS'23)  & 1.43 & 0.7s, 200 &4.60 & 16.2s, 200 & 8.56 & 1.8m, 100 & 23.46 & 18.2m, 5 & 35.33 & 2.3h, 1 &12.30 \\
    & BQ (NeurIPS'23) & 5.33 &1.5s, 200 & 6.66 & 1.3m, 200 & 9.43 & 9.3m, 100 & 27.65 & 1.9h, 5 & 41.80 & 13.5h, 1 & 15.21 \\
    & ELG (IJCAI'24) & 2.67 & 0.5s, 200 & 11.31 & 4.2s, 200 & 15.27 & 15.6s, 200 & 25.73 & 40.5s, 5 & 31.01 & 3.7m, 2 & 14.34 \\
    & INViT-3V (ICML'24) & 2.29 & 15.2s, 200 & 5.21 & 1.5m, 200 & 6.03 & 3.1m, 200 & 7.17 & 4.4m, 20 & 6.31 & 6.5m, 20 & 4.49 \\
    & GD (TMLR'24) & 2.29 & 3.1s, 200 & 6.87 & 36.4s, 200 & 25.26 & 3.2m, 200 & 329.10 & 26.7m, 20 & 627.83 & 3.4h, 4 & 198.41 \\
    & UDC$^\S$ (NeurIPS'24)  & 2.54 & 8.7s, 200 & 5.89 & 28.6s, 100 & 8.26 & 57.2s, 100 & 15.19 & 29.5s, 20 & 15.41 & 2.4m, 1 & 9.46 \\
    \cmidrule{2-13}
    & GELD + G (Ours)  & 3.28 & 0.6s, 200 & 4.41 & \textbf{1.8s}, 200 & 5.93 & \textbf{3.6s}, 200 & 11.62 & \textbf{10.8s}, 20 & 12.53 & \textbf{21.6s}, 20 & 7.55 \\
    & GELD + S$^*$ (Ours)  & \textbf{0.46} & 19.2s, 200 & \textbf{1.23} & 1.6m, 200 & \textbf{2.24} & 3.7m, 200 & \textbf{4.27} & 1.8m, 20 & \textbf{3.44} & 3.9m, 20 & \textbf{2.33} \\
    \cmidrule{1-13}
    \multirow{13}{*}{\rotatebox{90}{\textbf{explosion}}}& LKH3 & - & 2.7m, 1 & - & 3.8h, 1 & - & 15.6h, 1 & - & 1.7h, 1 & - & 1.3d, 1 & -\\
    \cmidrule{2-13}
    & Omni-TSP (ICML'23) & 2.05 & \textbf{0.3s}, 200 & 9.25 & 9.6s, 200 & 19.95 & 1.2m, 100 & 51.28 & 16.1m, 5 & 65.37 & 2.0h, 1 &24.69 \\
    & LEHD (NeurIPS'23) & 0.63 & 0.7s, 200 & 2.65 & 16.2s, 200 & 5.76 & 1.8m, 100 & 21.07 & 18.2m, 5 & 30.55 & 2.3h, 1 &10.12 \\
    & BQ (NeurIPS'23) & 5.97 &1.5s, 200 & 4.88 & 1.3m, 200 & 7.11 & 9.3m, 100 & 29.39 & 1.9h, 5 & 51.54 & 13.5h, 1 & 16.41 \\
    & ELG (IJCAI'24) & 0.87 & 0.5s, 200 & 9.27 & 4.2s, 200 & 13.67 & 15.6s, 200 & 22.79 & 40.5s, 5 & 23.46 & 3.7m, 2 & 11.68 \\
    & INViT-3V (ICML'24) & 1.62 & 15.2s, 200 & 5.54 & 1.5m, 200 & 7.32 & 3.1m, 200 & 9.92 & 4.4m, 20 & 7.85 & 6.5m, 20 & 5.37 \\
    & GD (TMLR'24) & 0.68 & 3.1s, 200 & 3.32 & 36.4s, 200 & 12.33 & 3.2m, 200 & 271.55 & 26.7m, 20 & 682.40 & 3.4h, 4 & 194.07 \\
    & UDC$^\S$ (NeurIPS'24)  & 0.66 & 8.6s, 200 & 4.60 & 28.6s, 200 & 6.96 & 57.2s, 100 & 16.15 & 29.5s, 20 & 17.44 & 2.4m, 1 & 9.16 \\
    \cmidrule{2-13}
    & GELD + G (Ours)  & 1.67 & 0.6s, 200 & 3.79 & \textbf{1.8s}, 200 & 5.40 & \textbf{3.6s}, 200 & 12.13 & \textbf{10.8s}, 20 & 14.27 & \textbf{21.6s}, 20 & 7.45 \\
    & GELD + S$^*$ (Ours)  & \textbf{0.18} & 19.2s, 200 & \textbf{0.95} & 1.6m, 200 & \textbf{1.52} & 3.7m, 200 & \textbf{4.55} & 1.8m, 20 & \textbf{4.70} & 3.9m, 20 & \textbf{2.39} \\
    \cmidrule{1-13}
    \multirow{13}{*}{\rotatebox{90}{\textbf{implosion}}}& LKH3 & - & 2.6m, 1 & - & 3.6h, 1 & - & 15.2h, 1 & - & 2.2h, 1 & - & 1.3d, 1 & -\\
    \cmidrule{2-13}
    & Omni-TSP (ICML'23) & 2.04 & \textbf{0.3s}, 200 & 8.63 & 9.6s, 200 & 19.18 & 1.2m, 100 & 50.37 & 16.1m, 5 & 62.58 & 2.0h, 1 & 23.83 \\
    & LEHD (NeurIPS'23) & 1.13 & 0.7s, 200 & 2.57 & 16.2s, 200 & 4.10 & 1.8m, 100 & 17.48 & 18.2m, 5 & 26.46 & 2.3h, 1 & 8.62 \\
    & BQ (NeurIPS'23) & 5.44 &1.5s, 200 & 4.84 & 1.3m, 200 & 5.22 & 9.3m, 100 & 16.42 & 1.9h, 5 & 25.23 & 13.5h, 1 & 9.56 \\
    & ELG (IJCAI'24) & 0.91 & 0.5s, 200 & 8.44 & 4.2s, 200 & 12.40 & 15.6s, 200 & 18.95 & 40.5s, 5 & 18.73 & 3.7m, 2 & 9.89 \\
    & INViT-3V (ICML'24) & 1.79 & 15.2s, 200 & 4.84 & 1.5m, 200 & 5.64 & 3.1m, 200 & 6.85 & 4.4m, 20 & 5.41 & 6.5m, 20 & 4.07 \\
    & GD (TMLR'24) & 1.45 & 3.1s, 200 & 4.29 & 36.4, 200 & 8.68 & 3.2m, 200 & 100.05 & 26.7m, 20 & 259.46 & 3.4h, 4 & 74.74 \\
    & UDC$^\S$ (NeurIPS'24)  & 0.54 & 8.7s, 200 & 3.29 & 28.7s, 200 & 3.74 & 57.2s, 100 & 7.74 & 29.5s, 20 & 10.04 & 2.4m, 1 & 5.07 \\
    
    \cmidrule{2-13}
    & GELD + G (Ours)  & 2.23 & 0.6s, 200 & 4.71 & \textbf{1.8s}, 200 & 4.98 & \textbf{3.6s}, 200 & 9.23 & \textbf{10.8s}, 20 & 10.02 & \textbf{21.6s}, 20 & 6.25 \\
    & GELD + S$^*$ (Ours)  & \textbf{0.22} & 19.2s, 200 & \textbf{1.29} & 1.6m, 200 & \textbf{1.64} & 3.7m, 200 & \textbf{3.14} & 1.8m, 20 & \textbf{2.81} & 3.9m, 20 & \textbf{1.84} \\
    \bottomrule
    \end{tabular}}
    \begin{tablenotes}
    \item The number in parentheses following ``TSP-$n$" indicates the total number of TSP-$n$ test instances. Symbol ``$\S$" indicates the model adopts a D\&C strategy.
    \end{tablenotes}
\end{table*}

\begin{table*}[!t]
\small
  \centering
    \caption{Performance comparisons on real-world TSPLIB95 and National TSP (Map of Countries) instances} \label{Tab_real}
    \vspace{-5pt}
    \resizebox{2\columnwidth}{!}{
    \begin{tabular}{l|l|c|c|c|c|c|c|c}
    \toprule
   \multicolumn{2}{c|}{\textbf{TSP-\{set\}}} & \textbf{$<$101} & \textbf{101-500} & \textbf{501-1000} & \textbf{1001-5000} & \textbf{5001-10000} & \textbf{$>$10000} & \textbf{(Total) gap$\downarrow$, time$\downarrow$} \\
    \cmidrule{1-9}
    \multirow{10}{*}{\rotatebox{90}{\textbf{TSPLIB95}}} & Total number of instances & 12 & 30 & 6 & 22 & 2 & 5 & 77\\
    \cmidrule{2-9}
    & Omni-TSP (ICML'23) & 6.87\% & 8.79\% & 19.59\% & 32.31\% & 63.28\% & OOM & \underline{(72)} 18.07\%, 3.8s \\
    &  LEHD (NeurIPS'23) & 0.61\% & 2.96\% & 4.05\% & 11.27\% & 24.14\% & \underline{(3)} 50.21\% & \underline{(75)} 7.56\%, 47.0s \\
    &  BQ (NeurIPS'23) & 8.64\% & 8.40\% & 8.08\% & 13.33\% & 27.37\% & \underline{(1)} 45.21\% & \underline{(73)} 10.92\%, 1.4m \\
    &  ELG (IJCAI'24) & 1.56\% & 4.55\% & 9.25\% & 12.61\% & 17.31\% & OOM & \underline{(72)} 7.25\%, 6.1s \\
    &  INViT-3V (ICML'24) & 1.15\% & 3.38\% & 6.33\% & 7.47\% & 9.34\% & 7.57\% & 4.86\%, 26.2s \\
    &  GD (TMLR'24) & 1.78\% & 4.29\% & 8.53\% & 52.17\% & 325.62\% & 991.24\% & 90.34\%,  2.7m\\
    &  UDC$^\S$ (NeurIPS'24) & \underline{(6)} 0.19\% & 2.18\% &  10.58\% & 13.00\% & 26.26\% & \underline{(1)} 23.37\% &\underline{(67)} 7.34\%,  5.6s\\
    \cmidrule{2-9}
    &  GELD (Ours) & 0.89\% & 4.92\% & 4.43\% & 8.91\% & 11.76\% & 15.90\% & 6.28\%, 3.8s \\
     &  GELD (Ours) +  BOTH& \textbf{0.26\%} & \textbf{1.56\%} & \textbf{1.92\%} & \textbf{3.44\%} & \textbf{7.09\%} & \textbf{5.96\%} & \textbf{2.35\%}, 27.6s \\
     \cmidrule{1-9}
    \multirow{10}{*}{\rotatebox{90}{\textbf{National TSPs}}} & Total number of instances & 2 & 1 & 3 & 4 & 9 & 8 & 27\\
    \cmidrule{2-9}
    & Omni-TSP (ICML'23) & 2.63\% & 10.44\% & 17.88\% & 71.65\% & 83.24\% & \underline{(1)} 71.67\% & \underline{(20)} 58.83\%, 2.3m \\
    &  LEHD (NeurIPS'23) & 0.12\% & 27.15\% & 44.20\% & 56.58\% & 93.92\% & \underline{(1)} 98.52\% & \underline{(20)} 66.51\%, 2.8m \\    
    &  BQ (NeurIPS'23) & 24.29\% & 12.18\% & 10.25\% & 40.55\% & 94.96\% & \underline{(1)} 55.65\% & \underline{(20)} 58.20\%, 14.8m \\
    &  ELG (IJCAI'24) & 2.28\% & 7.06\% & 12.55\% & 34.93\% & 48.95\% & (1) 22.44\% & \underline{(20)} 32.60\%, 3.7m \\
    &  INViT-3V (ICML'24) & 0.03\% & 2.88\% & 5.63\% & 10.17\% & 11.17\% & \underline{(7)} 9.48\% & \underline{(26)} 8.75\%, 3.4m \\
    &  GD (TMLR'24) + G & 3.51\% & 236.40\%& 921.61\% & 2093.71\% & 3868.87\% & \underline{(4)} 5236.23\% & \underline{(23)} 2919.47\% , 8.9m \\
    & UDC$^\S$ (NeurIPS'24) & - & 0.58\% &  10.04\% & 18.18\% &  25.44\% & \underline{(1)} 18.41\% &\underline{(18)} 19.49\%,  6.3s\\
    \cmidrule{2-9}
    &  GELD + G (Ours) & 0.41\% & 0.53\% & 5.10\% & 14.80\% & 17.99\% & 18.80\% &  14.39\%, 23.4s \\
     &  GELD + S$^*$ (Ours) & \textbf{0.02\%} & \textbf{0.02\%} & \textbf{2.12\%} & \textbf{6.97\%} & \textbf{7.66\%} & \textbf{8.21\%} & \textbf{6.26\%}, 1.4m \\
     \bottomrule
     \end{tabular}}
         \begin{tablenotes}

    
    \item[1] For each model, we report the average gap and inference time for the instances it successfully solves within a given set in our computer. Symbol ``OOM" is used to indicate cases where the model fails to solve all instances in the set due to the GPU memory constraint. Symbol ``\underline{($i$)}" denotes the number of instances the model successfully solves in this set. The absence of these two symbols indicates that the model can solve all instances in the set. Moreover, UDC fails to solve instances with sizes smaller than 100 nodes due to unknown errors.
    \end{tablenotes}
\end{table*}

\subsection{GELD as a Post-processing Technique}
We apply GELD in combination with PRC(1,000) to assess its effectiveness as a post-processing method for improving the solution quality of baseline models. Because the baseline models struggle with certain large-scale instances (e.g., CH71009 with 71,009 nodes), we introduce a simple and generic heuristic---RI---as an additional baseline. RI greedily selects the insertion point for each node, minimizing the insertion cost. We use the National TSPs dataset as the benchmark and apply GELD to reconstruct the solution generated by these baselines.

The results, as presented in Table~\ref{Tab_post}, demonstrate that our model significantly improves the solution quality by \textbf{at least 35\%} with an affordable increase in computing time, thereby highlighting the efficacy of GELD as a post-processing method. Moreover, the successful integration with RI, characterized by low latency and high solution quality, suggests that combining GELD with heuristic algorithms is a promising approach for efficiently solving large-scale TSPs. 

To further demonstrate the effectiveness of combining GELD with heuristic algorithms, we conduct additional experiments on the four extremely large TSPs, with sizes ranging from 104,815 to 744,710 nodes. As shown in Table~\ref{Tab_large}, GELD efficiently solves these extremely large TSPs. To the best of our knowledge, \textbf{our proposed approach is the first neural model capable of solving TSPs with up to 744,710 nodes without relying on D\&C strategies}.

\begin{table*}[!t]
  \centering
    \caption{Performance of baselines on National TSPs using GELD as a post-processing method} \label{Tab_post}
    \resizebox{2\columnwidth}{!}{
    \begin{tabular}{l|c|c|c|c|c|c|c|c}
    \toprule
  \textbf{TSP-\{set\}} & \textbf{$<$101} & \textbf{101-500} & \textbf{501-1000} & \textbf{1001-5000} & \textbf{5001-10000} & \textbf{$>$10000} & \textbf{(Total) gap$\downarrow$, time$\downarrow$} &\textbf{Gain$\uparrow$}\\
  \cmidrule{1-9}
    Total number & 2 & 1 & 3 & 4 & 9 & 8 & \multicolumn{2}{c}{27}\\
        \cmidrule{1-9}
    Omni-TSP + GELD & 0.02\% & 0.67\% & 1.61\% & 5.49\% & 7.13\% & \underline{(1)} 5.40\% & \underline{(20)} 4.85\%, +26.6s & \textbf{91.76\%}\\
    LEHD + GELD & 0.02\% & 7.12\% & 3.24\% & 8.40\% & 9.30\% & \underline{(1)} 11.64\% & \underline{(20)} 7.29\%, +26.6s & \textbf{89.04\%} \\
   BQ + GELD & 0.02\% & 4.52\% & 3.48\% & 7.01\% & 9.59\% & \underline{(1)} 8.81\% & \underline{(20)} 6.91\%, +26.6s & \textbf{88.13\%} \\
    ELG + GELD & 2.15\% & 0.67\% & 2.56\% & 8.78\% & 17.00\% & \underline{(1)} 7.99\% & \underline{(20)} 10.44\%, +26.6s & \textbf{67.98\%}\\
    INViT-3V + GELD & 0.02\% & 0.68\% & 2.24\% & 4.57\% & 5.35\% & \underline{(7)} 4.74\% & \underline{(26)} 4.12\%, +34.0s & \textbf{52.91\%}\\
    GD + GELD & 0.29\% & 4.52\% & 22.66\% &  65.78\% & 131.14\% & \underline{(4)} 140.89\% & \underline{(23)} 90.43\%, +28.3s & \textbf{96.90\%} \\
    UDC + GELD & - & 0.58\% & 7.59\% &  11.30\% & 16.45\% & \underline{(1)} 11.19\% & \underline{(18)} 12.66\%, +26.2s & \textbf{35.05\%} \\
    \cmidrule{1-9}
    RI & 8.77\% & 11.54\% & 11.53\% & 12.49\% & 13.48\% & 13.34\% & 12.66\%, 1.1s  &\multirow{2}{*}{\textbf{78.59\%}}\\
    \textbf{+} GELD & 0.02\% & 2.35\% & 1.75\% & 3.14\% & 3.29\% & 2.90\% & \textbf{2.71\%}, +36.8s \\
     \bottomrule
     \end{tabular}}
    \begin{tablenotes}
    \item[1] Gain is calculated as 1-(the result of baseline with GELD)/(the result of baseline without GELD).
    \end{tablenotes}
\end{table*}

\begin{table*}[!t]
  \centering
    \caption{Ablation studies on synthetic TSP instances of the uniform distribution} 
    \label{Tab_abla}
    \resizebox{2.1\columnwidth}{!}{
    \begin{tabular}{ll|cc|cc|cc|cc|cc}
    \toprule
    \multicolumn{2}{c|}{\multirow{2}{*}{Model + Inference}} & \multicolumn{2}{c|}{\textbf{TSP-100 (200)}} & \multicolumn{2}{c}{\textbf{TSP-500 (200)}} & \multicolumn{2}{c|}{\textbf{TSP-1000 (200)}} & \multicolumn{2}{c|}{\textbf{TSP-5000 (20)}} & \multicolumn{2}{c}{\textbf{TSP-10000 (20)}}\\
    & & gap(\%)$\downarrow$ & time$\downarrow$, $n_{\textit{bs}}\uparrow$ & gap(\%)$\downarrow$ & time$\downarrow$, $n_{\textit{bs}}\uparrow$ & gap(\%)$\downarrow$ & time$\downarrow$, $n_{\textit{bs}}\uparrow$ & gap(\%)$\downarrow$ & time$\downarrow$, $n_{\textit{bs}}\uparrow$ & gap(\%)$\downarrow$ & time$\downarrow$, $n_{\textit{bs}}\uparrow$ \\
    \cmidrule{1-12}
    \multirow{3}{*}{w/o RALA} & G & 1.12 & 0.8s, 200 & 2.61 & 2.0s, 200 & 3.63 & 4.1s, 200 & 11.67 & 45.2s, 5 & 12.48 & 3.7m, 2\\
    & S$^*$ & 0.05 & 20.1s, 200 & 0.48 & 1.7m, 200 & 0.64 & 3.6m, 200 & 4.18 & 3.9m, 5 & 4.04 & 27.3m, 1\\
    & - Norm & 0.05 & 20.1s, 200 & 0.50 & 1.7m, 200 & 0.72 & 3.6m, 200 & 5.25 & 3.9m, 5 & 5.76 & 27.3m, 1\\
    \cmidrule{1-12}
    \multirow{3}{*}{\tabincell{l}{w/o second\\stage training}} & G & 0.86 & 0.6s, 200 & 3.28 & 1.8s, 200 & 4.17 & 3.6s, 200 & 13.61 & 10.8s, 20 & 15.21 & 21.6s, 20\\
    & S$^*$ & 0.05 & 19.2s, 200 & 0.69 & 1.6m, 200 & 1.14 & 3.7m, 200 & 3.73 & 1.8m, 20 & 3.10 & 3.9m, 20\\
    & - Norm & 0.05 & 19.2s, 200 & 0.74 & 1.6m, 200 & 1.39 & 3.7m, 200 & 5.50 & 1.8m, 20 & 5.62 & 3.9m, 20\\
    \cmidrule{1-12}
    \multirow{3}{*}{\tabincell{l}{w/o global\\view}} & G & 1.33 & 0.5s, 200 & 3.03 & 1.5s, 200 & 3.79 & 3.2s, 200 & 10.14 & 9.9s, 20 & 11.13 & 20.2s, 20\\
    & S$^*$ & 0.06 & 18.5s, 200 & 0.54 & 1.6m, 200 & 0.65 & 3.5m, 200 & 3.08 & 1.8m, 20 & 3.41 & 3.8m, 20\\
    & - Norm & 0.06 & 18.5s, 200 & 0.54 & 1.6m, 200 & 0.67 & 3.5m, 200 & 3.84 & 1.8m, 20 & 4.78 & 3.8m, 20\\
    \cmidrule{1-12}
    \multirow{3}{*}{\tabincell{l}{w/o local\\view}} & G & 1.32 & 0.6s, 200 & 2.13 & 6.1s, 200 & 2.51 & 33.6s, 200 & 4.82 & 4.1m, 20 & 5.57 & 30.9m, 5\\
    & S$^*$ & 0.08 & 19.2s, 200 & 0.44 & 3.0m, 100 & 0.42 & 12.6m, 40 & 1.92 & 1.4h, 1 & \multicolumn{2}{c}{\multirow{2}{*}{OOM}}\\
    & - Norm & 0.09 & 19.2s, 200 & 0.45 & 3.0m, 100 & 0.47 & 12.6m, 40 & 2.28 & 1.4h, 1 \\
        \cmidrule{1-12}
    \multirow{3}{*}{GELD} & G & 1.11 & 0.6s, 200 & 2.39 & 1.8s, 200 & 2.94 & 3.6s, 200 & 7.62 & 10.8s, 20 & 9.33 & 21.6s, 20\\
    & S$^*$  & 0.06 & 19.2s, 200 & 0.52 & 1.6m, 200 & 0.58 & 3.7m, 200 & 2.77 & 1.8m, 20 & 2.38 & 3.9m, 20\\    
    & - Norm  & 0.06 & 19.2s, 200 & 0.53 & 1.6m, 200 & 0.61 & 3.7m, 200 & 3.29 & 1.8m, 20 & 3.64 & 3.9m, 20\\   
     \bottomrule
     \end{tabular}}
         \begin{tablenotes}
    \item[1] Symbol ``- Norm" denotes without the normalization operation during the RC process.
    \end{tablenotes}
\end{table*}

\begin{table}[!t]
  \centering
    \caption{Performance of GELD on extremely large TSPs}\label{Tab_large}
     \resizebox{1\columnwidth}{!}{\begin{tabular}{lcccc}
    \toprule
    \multirow{2}{*}{Instances} & \multicolumn{2}{c}{\textbf{RI}} & \multicolumn{2}{c}{\textbf{+GELD}} \\
    & gap$\downarrow$ & time & gap$\downarrow$ (gain$\uparrow$)  & time \\
    \midrule
    sra104815 & 21.26\% & 52.2s & 9.67\% (54.66\%) & +2.7m \\
    ara238025 & 20.65\% & 5.4m  & 9.25\% (55.21\%) & +5.9m \\
    lra498378 & 18.94\% & 30.1m & 6.58\% (65.26\%) & +12.9m \\
    lrb744710 & 21.08\% & 1.7h & 8.97\% (57.45\%) & +19.7m \\
     \bottomrule
     \end{tabular}}
         \begin{tablenotes}
    \item[1] Other baseline models are unable to solve these extremely large TSPs due to the limitations of GPU memory.
    \end{tablenotes}
\end{table}

\subsection{Ablation Studies on GELD Design Choices}\label{Sec_abl}
We conduct extensive ablation studies to assess the effectiveness of key design choices in GELD, by investigating the following four aspects: 1)~The efficacy of RALA by replacing it with the standard attention mechanism \citep{Vaswani2017}; 2)~The impact of omitting the second-stage training and relying solely on the first-stage training; 3)~The role of the global view by removing the global information transfer module from GE; and 4)~The contribution of the local view by extending LD's local view to a global view, i.e., considering all available nodes as the candidate set (see the context of (\ref{eq11})). We retrain GELD under each of these modified conditions and evaluate its performance on synthetic TSPs of the uniform distribution. Moreover, during the testing phase, we assess the impact of diversifying model inputs by eliminating the normalization operation in the RC process.

We present the ablation study results in Table~\ref{Tab_abla}. Firstly, while GELD with the standard attention mechanism performs comparably to GELD with RALA on small-scale instances (TSP-\{100, 500\}), it experiences a performance degradation (especially in inference speed and parallel processing capability) on large-scale instances (TSP-\{5000, 10000\}). This finding demonstrates that RALA is critical for enabling GELD to solve TSPs in a short time period and scale effectively to larger-size instances. Secondly, incorporating the second-stage training leads to a 39.1\% improvement in solution quality compared to only applying the first-stage training, underscoring the importance of the two-stage training strategy. Notably, even without the second-stage training, GELD achieves an average gap of 1.74\%, outperforming all seven baseline models (see Table~\ref{tab_syn}). Thirdly, integrating the global view into GELD improves the average gap by 31.6\% when compared to using a local view only (i.e., removing the global information transfer module from GE), demonstrating the benefit of exploiting global information. Fourthly, while extending LD’s local view to a global view (i.e., considering all available nodes as candidates instead of set~$K_\textit{set}$) enhances solution quality, it significantly hampers inference speed and parallel processing capability, particularly in large-scale instances (TSP-\{10000\}). These results highlight the effectiveness of the local view in enabling GELD to efficiently solve TSPs across different scales. Last but not least, removing the normalization operation in the RC process deteriorates model performance in all aspects.

Furthermore, we conduct ablation studies to evaluate the sensitivity of two hyperparameters: the number of regions $m$ and the range of local selection $k_m$. To examine the impact of $m$, we use the first-stage training for GELD with three configurations, testing its effect on model performance across various TSP sizes and distributions: 1)~$m_r=m_c$=2, resulting in $m=4$ regions; 2) $m_r=m_c$=3, resulting in $m=9$ regions; and 3) $m_r=m_c$=4, resulting in $m=16$ regions. The results presented in Table~\ref{tab_m} indicate that increasing $m$ generally improves model performance albeit with a slight reduction in inference speed. Additionally, the variations in $m$ have an insignificant impact on the overall performance, demonstrating the model's robustness across different configurations.

For the range of local selection $k$, we test three values (namely 50, 100, 150) on the synthetic TSP instances of the uniform distribution. As shown in Table~\ref{Tab_k}, larger values of $k$ improve model performance while decreasing inference speed. These findings further highlight the importance of adopting LD in our model design.

\begin{table*}[!t]
  \centering
    \caption{Performance of the first-stage trained GELD with different $m$}\label{tab_m}
    \resizebox{2\columnwidth}{!}{
    \begin{tabular}{l|l|l|cc|cc|cc|cc|cc|c}
    \toprule
     & \multirow{2}{*}{\textbf{Value}} & \multirow{2}{*}{\textbf{Inference}} & \multicolumn{2}{c|}{\textbf{TSP-100}} & \multicolumn{2}{c}{\textbf{TSP-500}} & \multicolumn{2}{c|}{\textbf{TSP-1000}} & \multicolumn{2}{c|}{\textbf{TSP-5000}} & \multicolumn{2}{c|}{\textbf{TSP-10000}} & \multirow{2}{*}{\tabincell{l}{\textbf{Average}\\ \textbf{gap(\%)}$\downarrow$}
     }\\
    & & & gap(\%)$\downarrow$ & time$\downarrow$ & gap(\%)$\downarrow$ & time$\downarrow$ & gap(\%)$\downarrow$ & time$\downarrow$ & gap(\%)$\downarrow$ & time$\downarrow$ & gap(\%)$\downarrow$ & time$\downarrow$ \\
    \cmidrule{1-14}
    \multirow{6}{*}{\rotatebox{90}{\textbf{uniform}}}&  \multirow{2}{*}{4} & G & 0.71 & 0.6s & 3.81 & 1.7s & 4.81 & 3.4s & 15.75 & 10.3s & 17.08 & 20.8s & 8.43\\
    & & BOTH & 0.05 & 19.0s &	0.75 &	1.5m &	1.23 &3.6m & 4.1 & 1.7m & 3.69 & 3.8m & 1.96 \\
    \cmidrule{2-14}
    &  \multirow{2}{*}{9} & G & 0.86  & 0.6s & 3.28 & 1.8s & 4.17 & 3.6s & 13.61 & 10.8s & 15.21 & 21.6s & 7.43\\ 
    & & BOTH & 0.05 & 19.2s & 0.69 & 1.6m & 1.14 & 3.7m & 3.73 & 1.8m & 3.1 & 3.9m & 1.74\\ 
    \cmidrule{2-14}
    &  \multirow{2}{*}{16} & G & 0.94 & 0.6s & 3.44 & 1.9s & 5.15 & 3.9s & 12.68 & 11.4s & 12.15 & 23.4s & 6.87\\ 
    & & BOTH & 0.05 & 20.4s & 0.78 & 1.7m & 1.35 & 3.8m & 3.86 & 2.0m & 2.87 & 4.1m & 1.78\\ 

    \cmidrule{1-14}
    \multirow{6}{*}{\rotatebox{90}{\textbf{clustered}}}&  \multirow{2}{*}{4} & G & 3.49 & 0.6s & 7.4 & 1.7s & 10.83 & 3.4s & 20.59 & 10.3s & 32.15 & 20.8s & 14.89\\
    & & BOTH & 0.51 & 19.0s & 1.86 & 1.5m & 3.20 & 3.6m & 5.81 & 1.7m & 5.48 & 3.8m & 3.37\\ 
    \cmidrule{2-14}
    &  \multirow{2}{*}{9} & G & 2.80 & 0.6s & 6.73 & 1.8s & 10.34 & 3.6s & 16.36 & 10.8s & 15.31 & 21.6s & 10.31\\ 
    & & BOTH & 0.49 & 19.2s & 1.87 & 1.6m & 3.24 & 3.7m & 5.42 & 1.8m & 4.01 & 3.9m & 3.01\\ 
    \cmidrule{2-14}
    &  \multirow{2}{*}{16} & G & 2.65 & 0.6s & 6.9 & 1.9s & 9.76 & 3.9s & 14.63 & 11.4s & 16.06 & 23.4s &10.00\\
    & & BOTH & 0.42 & 20.4s & 1.66 & 1.7m & 2.79 & 3.8m & 5.36 & 2.0m & 4.23 & 4.1m & 2.89\\ 

    \cmidrule{1-14}
    \multirow{6}{*}{\rotatebox{90}{\textbf{explosion}}}&  \multirow{2}{*}{4} & G & 1.28 & 0.6s & 5.52 & 1.7s & 9.37 & 3.4s & 18.79 &10.3s & 28.56 & 20.8s & 12.70\\ 
    & & BOTH & 0.10 & 19.0s & 1.52 & 1.5m & 2.35 & 3.6m & 6.28 & 1.7m & 5.31 & 3.8m & 3.11\\ 
    \cmidrule{2-14}
    &  \multirow{2}{*}{9} & G & 1.00 & 0.6s & 4.68 & 1.8s & 8.02 & 3.6s & 17.11 & 10.8s & 17.87 & 21.6s & 9.74\\
    & & BOTH & 0.13 & 19.2s & 1.25 & 1.6m & 2.09 & 3.7m & 5.57 & 1.8m & 5.01 & 3.9m & 2.81 \\
    \cmidrule{2-14} 
    &  \multirow{2}{*}{16} & G & 1.34 & 0.6s & 5.9 & 1.9s & 9.59 & 3.9s & 14.71 & 11.4s & 17.51 & 23.4s & 9.81\\ 
    & & BOTH & 0.27 & 20.4s & 1.5 & 1.7m & 2.5 & 3.8m & 5.55 & 2.0m & 5.05 & 4.1m & 2.97\\ 

    \cmidrule{1-14}
    \multirow{6}{*}{\rotatebox{90}{\textbf{implosion}}}&  \multirow{2}{*}{4} & G & 1.85 & 0.6s & 5.47 & 1.7s & 8.07 & 3.4s & 18.1 & 10.3s & 21.36 & 20.8s & 10.97\\
    & & BOTH & 0.17 & 19.0s & 1.37 & 1.5m & 2.17 & 3.6m & 5.18 & 1.7m & 4.3 & 3.8m & 2.64\\ 
    \cmidrule{2-14}
    &  \multirow{2}{*}{9} & G & 1.84 & 0.6s & 5.2 & 1.8s & 6.61 & 3.6s & 14.84 & 10.8s & 15.69 & 21.6s & 8.84\\
    & & BOTH & 0.17 & 19.2s & 1.35 & 1.6m & 1.9 & 3.7m & 4.39 & 1.8m & 3.78 & 3.9m & 2.32\\ 
    \cmidrule{2-14} 
    &  \multirow{2}{*}{16} & G & 1.86 & 0.6s & 5.05 & 1.9s & 7.37 & 3.9s & 13.89 & 11.4s & 12.92 & 23.4s & 8.22\\
    & & BOTH & 0.23 & 20.4s & 1.44 & 1.7m & 2.23 & 3.8m & 4.62 & 2.0m & 3.27 & 4.1m & 2.36\\ 
    \bottomrule
    \end{tabular}}
\end{table*}

\begin{table*}[!t]
  \centering
    \caption{Performance of GELD with different $k_m$ on synthetic TSP instances of the uniform distribution} \label{Tab_k}
    \resizebox{2\columnwidth}{!}{
    \begin{tabular}{l|l|cc|cc|cc|cc|cc|c}
    \toprule
    \multicolumn{1}{c|}{\multirow{2}{*}{Value}} & \multirow{2}{*}{Inference} & \multicolumn{2}{c|}{\textbf{TSP-100}} & \multicolumn{2}{c|}{\textbf{TSP-500}} & \multicolumn{2}{c|}{\textbf{TSP-1000}} & \multicolumn{2}{c|}{\textbf{TSP-5000}} & \multicolumn{2}{c|}{\textbf{TSP-10000}} & \multirow{2}{*}{\tabincell{l}{\textbf{Average}\\ \textbf{gap(\%)}$\downarrow$}
     }\\
    
    & & gap(\%)$\downarrow$ & time$\downarrow$ & gap(\%)$\downarrow$ & time$\downarrow$ & gap(\%)$\downarrow$ & time$\downarrow$ & gap(\%)$\downarrow$ & time$\downarrow$ & gap(\%)$\downarrow$ & time$\downarrow$\\
    \cmidrule{1-13}
    \multirow{2}{*}{50} & G & 2.00 & 0.4s & 4.02 & 1.2s & 5.00 & 2.9s & 10.19 & 9.2s & 11.03 & 18.9s & 6.45\\
    & BOTH & 0.17 & 18.7s &  0.85 & 1.4m & 1.06 & 2.9m & 3.20 & 1.5m & 3.15 & 3.5m & 1.69\\
    \cmidrule{1-13}
    \multirow{2}{*}{100} & G & 1.11  & 0.6s & 2.39 & 1.8s & 2.94 & 3.6s & 7.62 & 10.8s & 9.33 & 21.6s & 4.68 \\
    & BOTH & 0.06 & 19.2s & 0.52 & 1.6m & 0.58 & 	3.7m & 2.77 & 1.8m & 2.38 & 3.9m &1.26 \\
    \cmidrule{1-13}
    \multirow{2}{*}{150} & G & 1.11 & 0.7s & 2.28 & 2.4s & 2.33 & 5.4s & 7.05 & 12.1s & 9.52 & 24.1s & 4.46\\
    & BOTH & 0.06 & 21.6s & 0.45 & 2.0m & 0.49 & 4.3m &	3.11 & 2.4m & 2.04 & 4.9m &1.23\\
     \bottomrule
     \end{tabular}}
\end{table*}

\subsection{Case Studies on Real-world Scenarios}
To better illustrate the effectiveness of applying GELD in real-world scenarios, we use two country maps from the National TSP dataset as case study instances. Specifically, we select TZ6117 and FI10639 as representatives of real-world TSP instances and compare the results of GELD, SOTA baseline models (LEHD \citep{Luo2023} and INViT-3V \citep{Fang2024}), and LKH3. LKH3 is widely recognized as the most efficient method for producing near-optimal TSP solutions. We use LKH3 with 20,000 iterations over 10 runs to solve these instances.  As shown in Table~\ref{tab:case}, GELD significantly reduces inference time for LKH3, achieving an average speedup of 36$\times$, while maintaining comparable tour lengths (an average drop of 4.47\%).  Compared to the baseline models, GELD outperforms them in both inference time (5$\times$ faster) and tour lengths (an average improvement of 37.83\%). These results further highlight GELD’s potential for practical applications in real-world scenarios.
\begin{table}[!t]
\centering
\caption{Case study results of GELD and baseline methods on real-world instances}
\label{tab:case}
\resizebox{1\columnwidth}{!}{
\begin{tabular}{l|cccccc}
\toprule
\multirow{2}{*}{\textbf{Method}} & \multicolumn{3}{c|}{\textbf{TZ6117}} &\multicolumn{3}{c}{\textbf{FI10639}} \\
& length$\downarrow$ &  gap$\downarrow$ & \multicolumn{1}{c|}{time$\downarrow$}   & length$\downarrow$ & gap$\downarrow$ & time$\downarrow$ \\

\midrule
LKH3&  394,629 & 0.01\%& \multicolumn{1}{c|}{27.0m} & 520,415 & 0.01\% &37.1m \\
LEHD & 596,806 & 51.24\% & \multicolumn{1}{c|}{2.6m} & 1,033,064 & 98.52\% & 3.1m \\
INViT-3V & 431,900 & 9.45\% & \multicolumn{1}{c|}{2.4m} & 572,630 & 10.04\% & 2.7m \\
\hline

GELD & 416,865 & 5.63\% & \multicolumn{1}{c|}{1.0m} & 554,364 &  6.53\%&  1.6m\\
RI + GELD & 405,500 & 2.76\% & \multicolumn{1}{c|}{34.8s} & 535,994 &  3.00\%&  48.0s\\
\bottomrule
\end{tabular}}
\end{table}

\begin{figure*}[!t]
\centering
\subfigure[Optimal solution]{\includegraphics[width=.8\columnwidth]{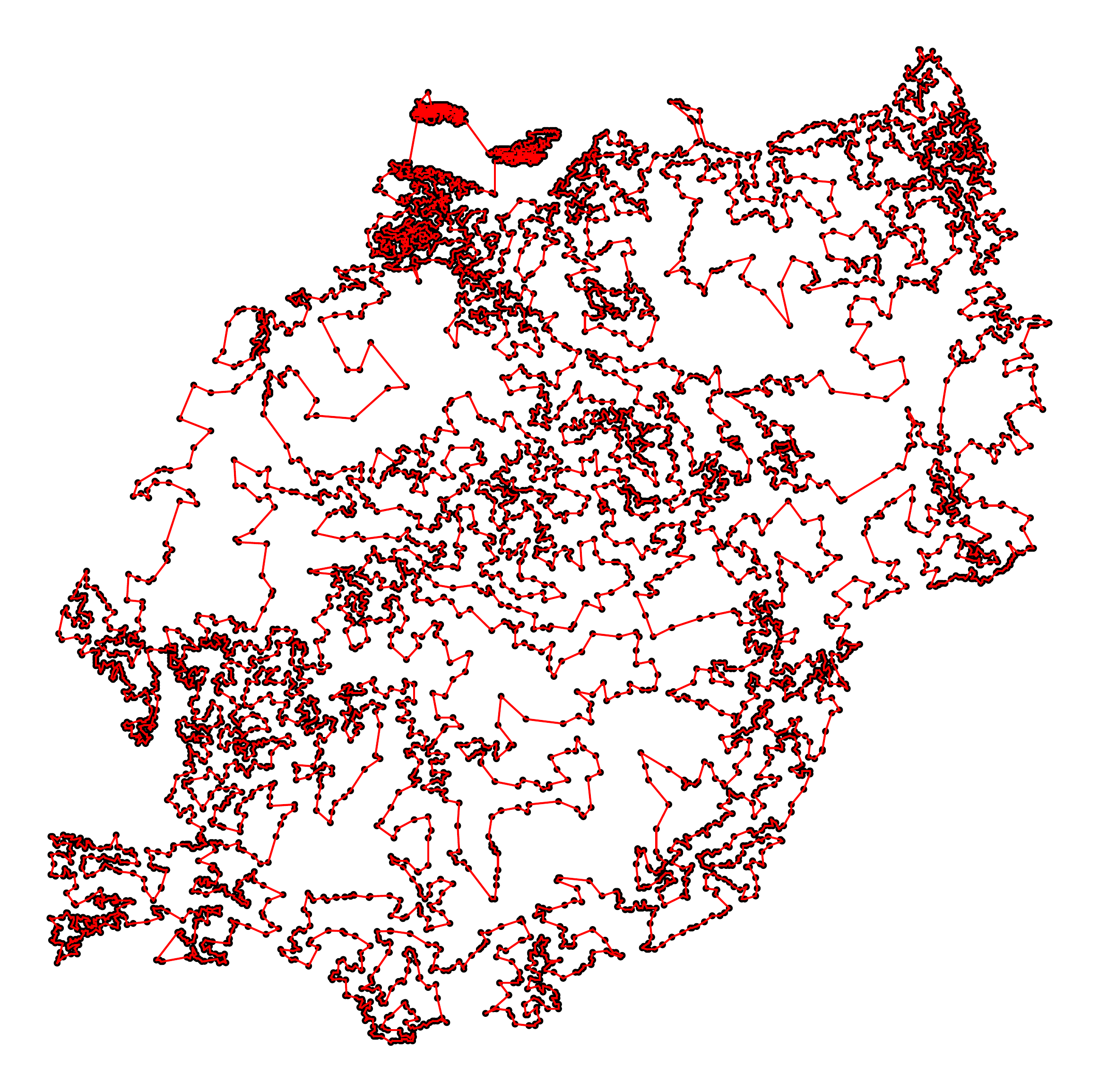}}
\subfigure[LEHD: gap=51.24\%]{\includegraphics[width=.8\columnwidth]{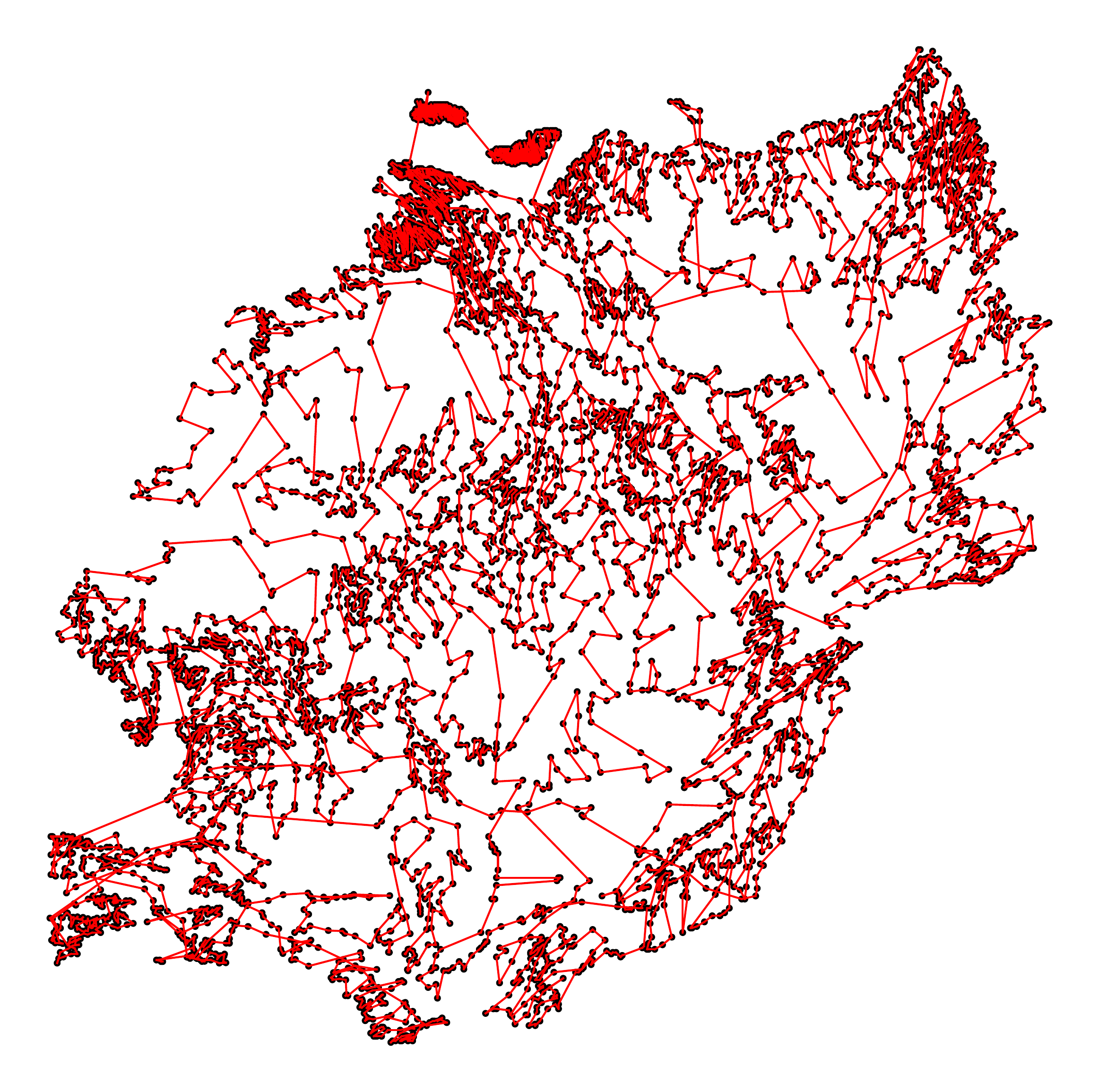}}
\subfigure[INViT: gap=9.45\%]{\includegraphics[width=.8\columnwidth]{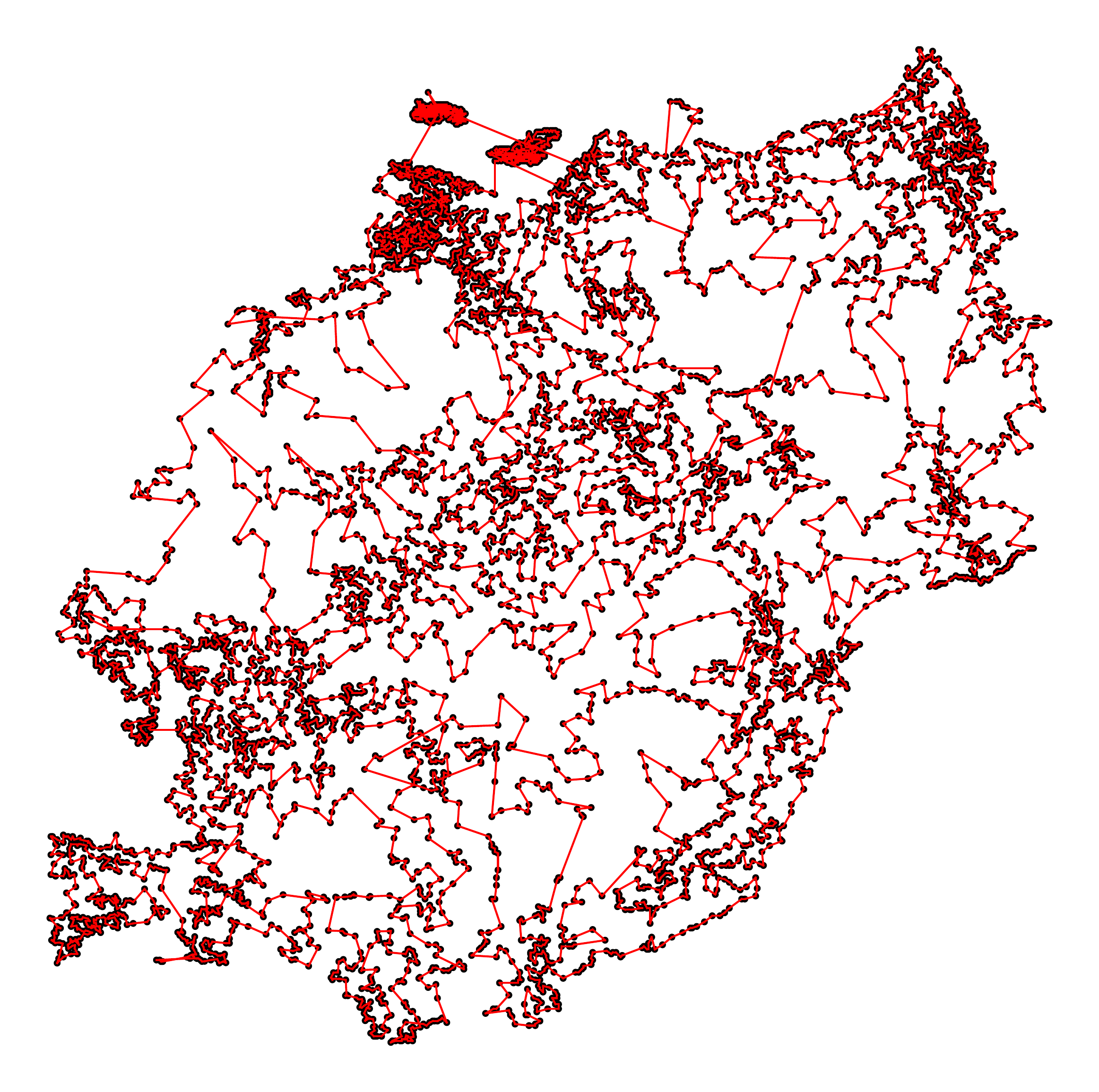}}
\subfigure[GELD (Ours): gap=5.63\%]{\includegraphics[width=.8\columnwidth]{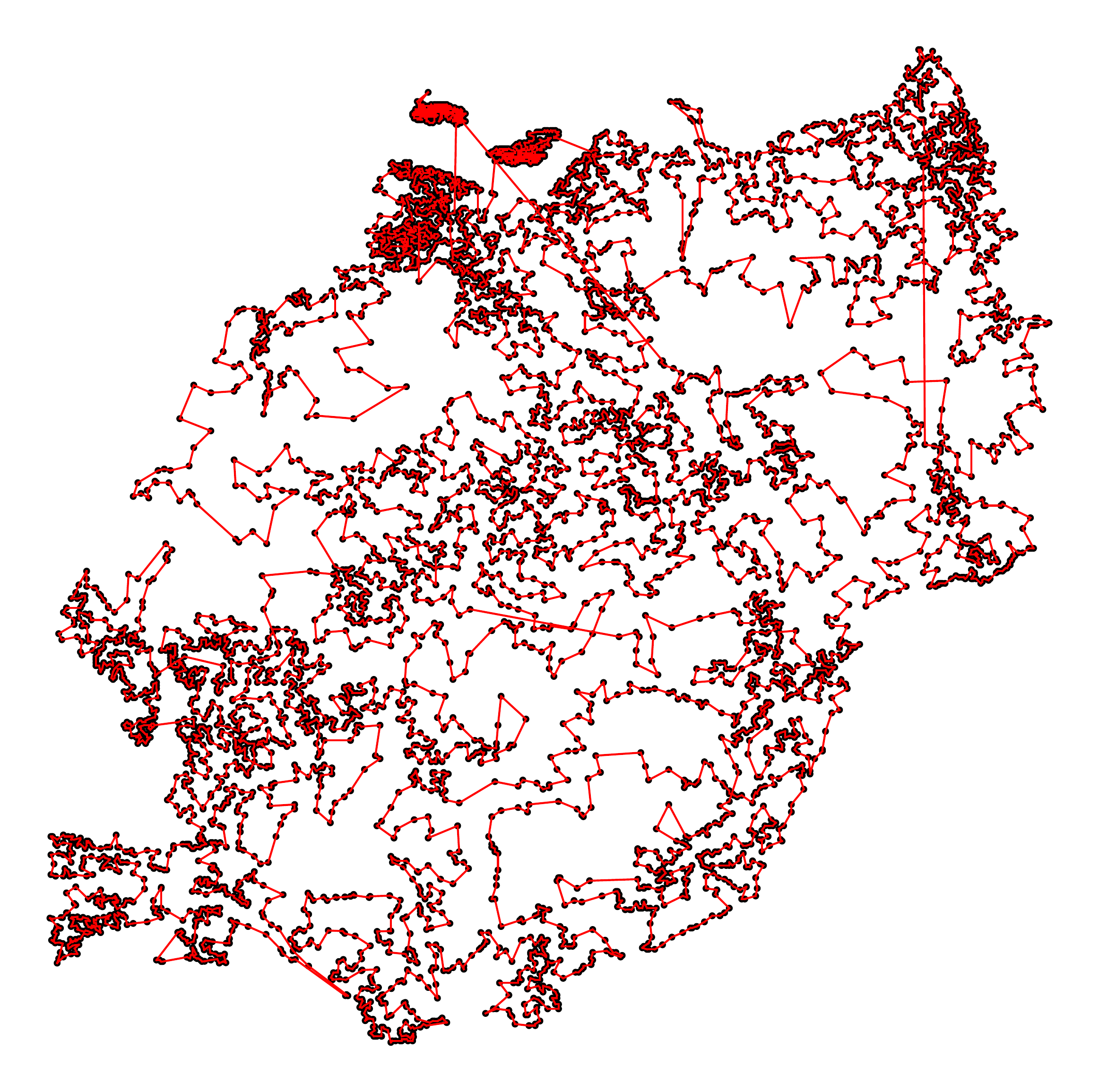}}
\subfigure[RI: gap=14.42\%]{\includegraphics[width=.8\columnwidth]{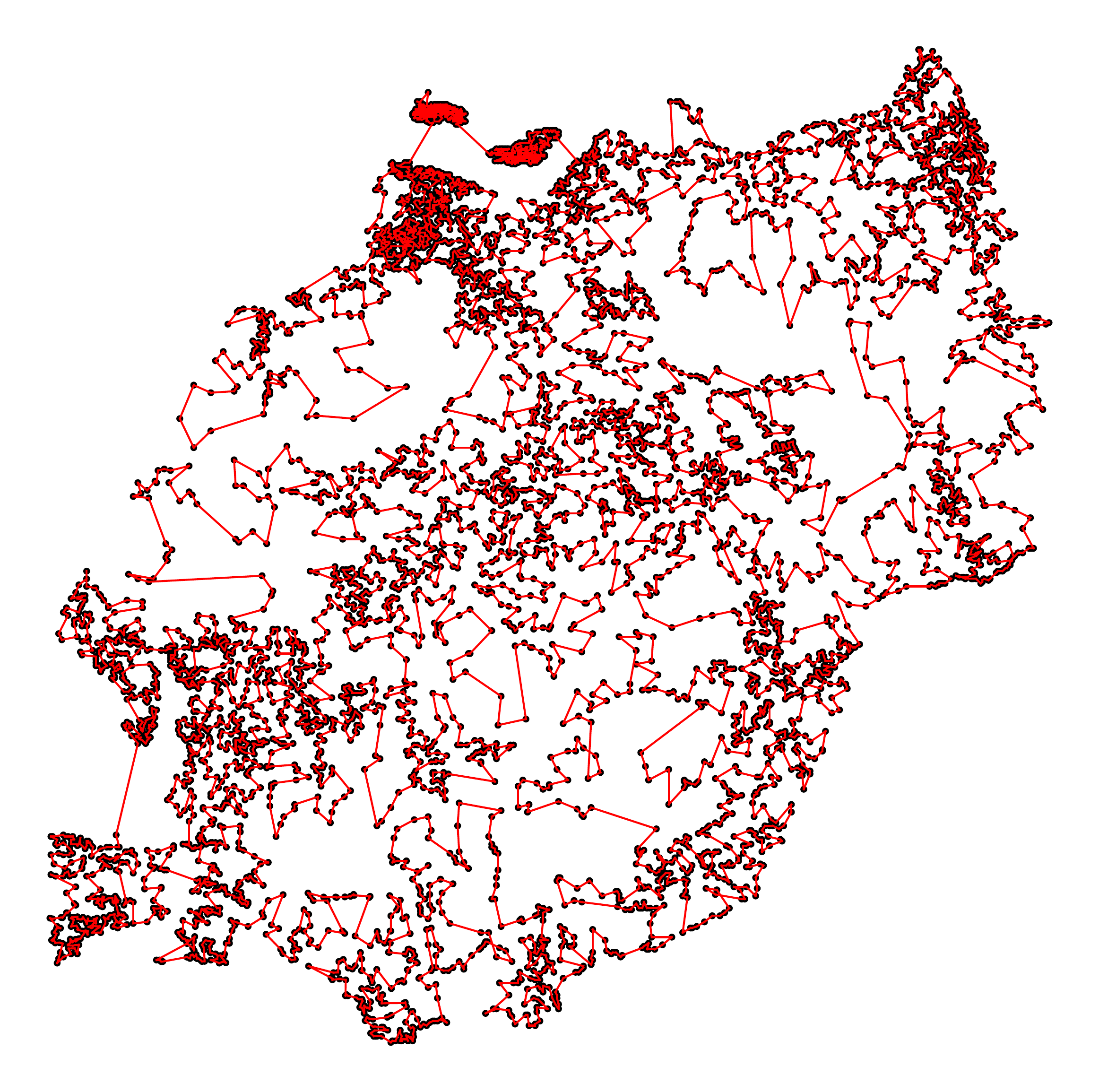}}
\subfigure[RI + GELD (Ours): gap=2.76\%]{\includegraphics[width=.8\columnwidth]{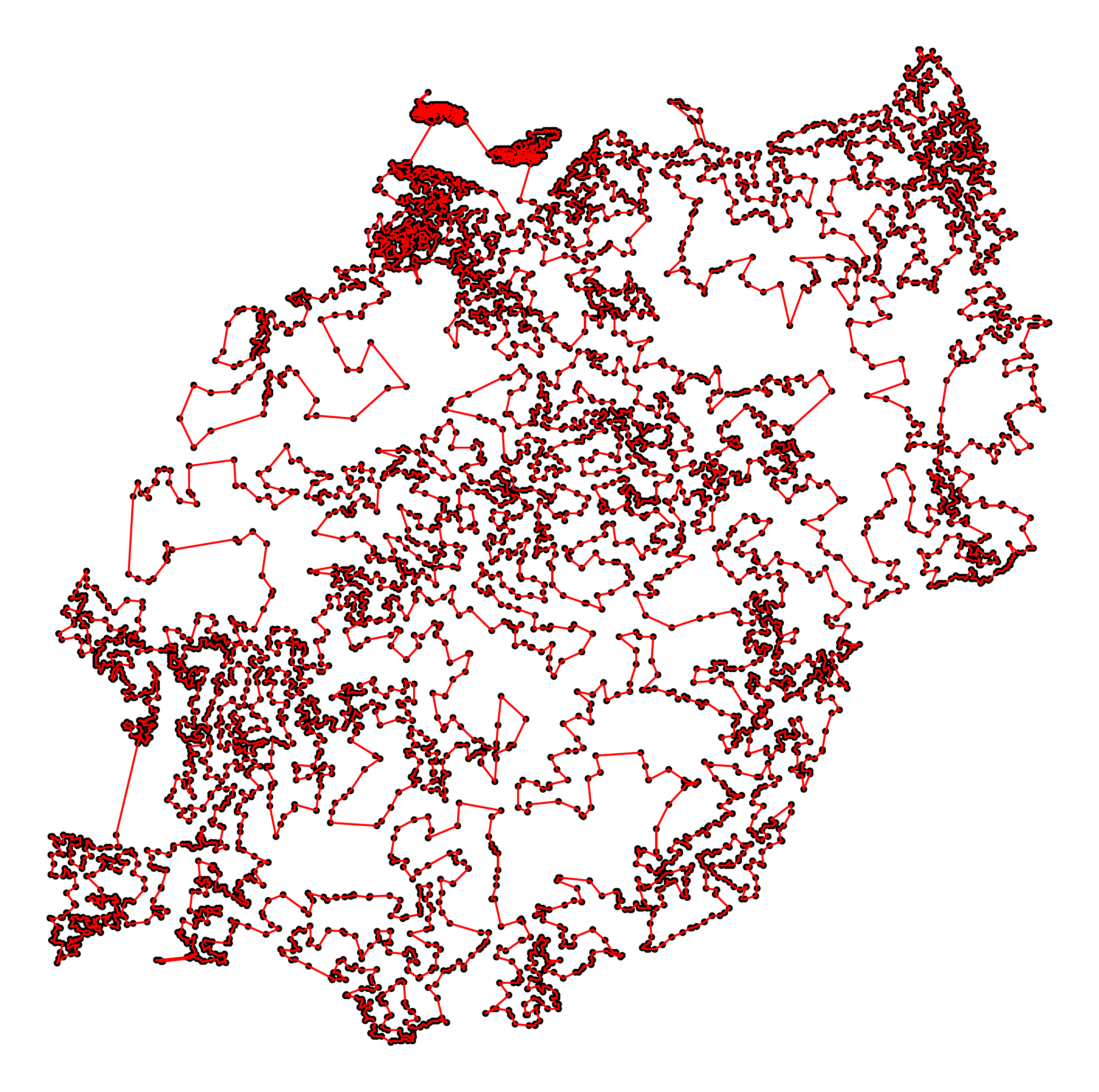}}
\caption{Visualization of solutions on TZ6117 TSP instance with 6117 nodes.}\label{fig_TZ}
\end{figure*}

\begin{figure*}[!t]
\centering
\subfigure[Optimal solution]{\includegraphics[width=.8\columnwidth]{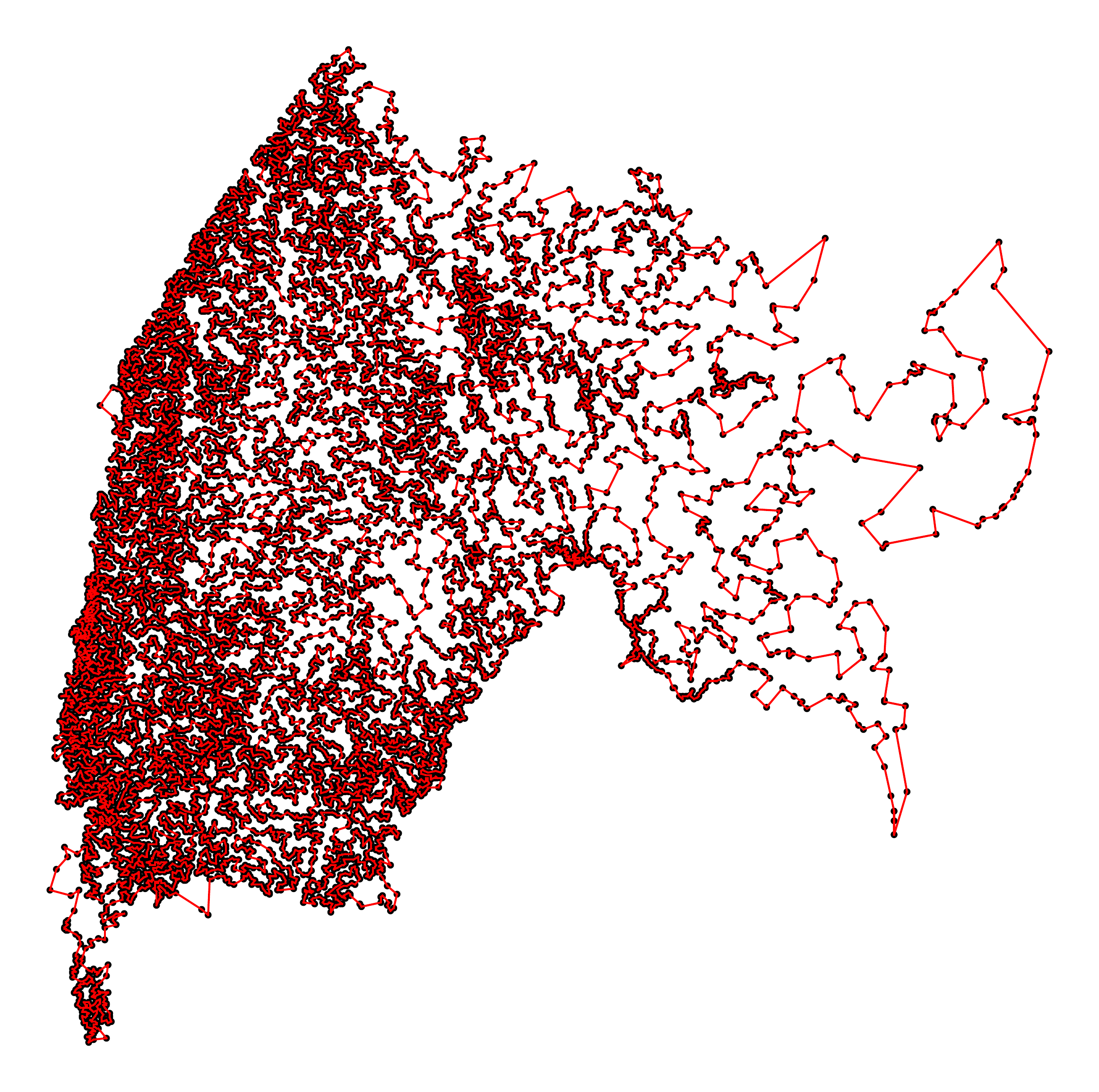}}
\subfigure[LEHD: gap=98.52\%]{\includegraphics[width=.8\columnwidth]{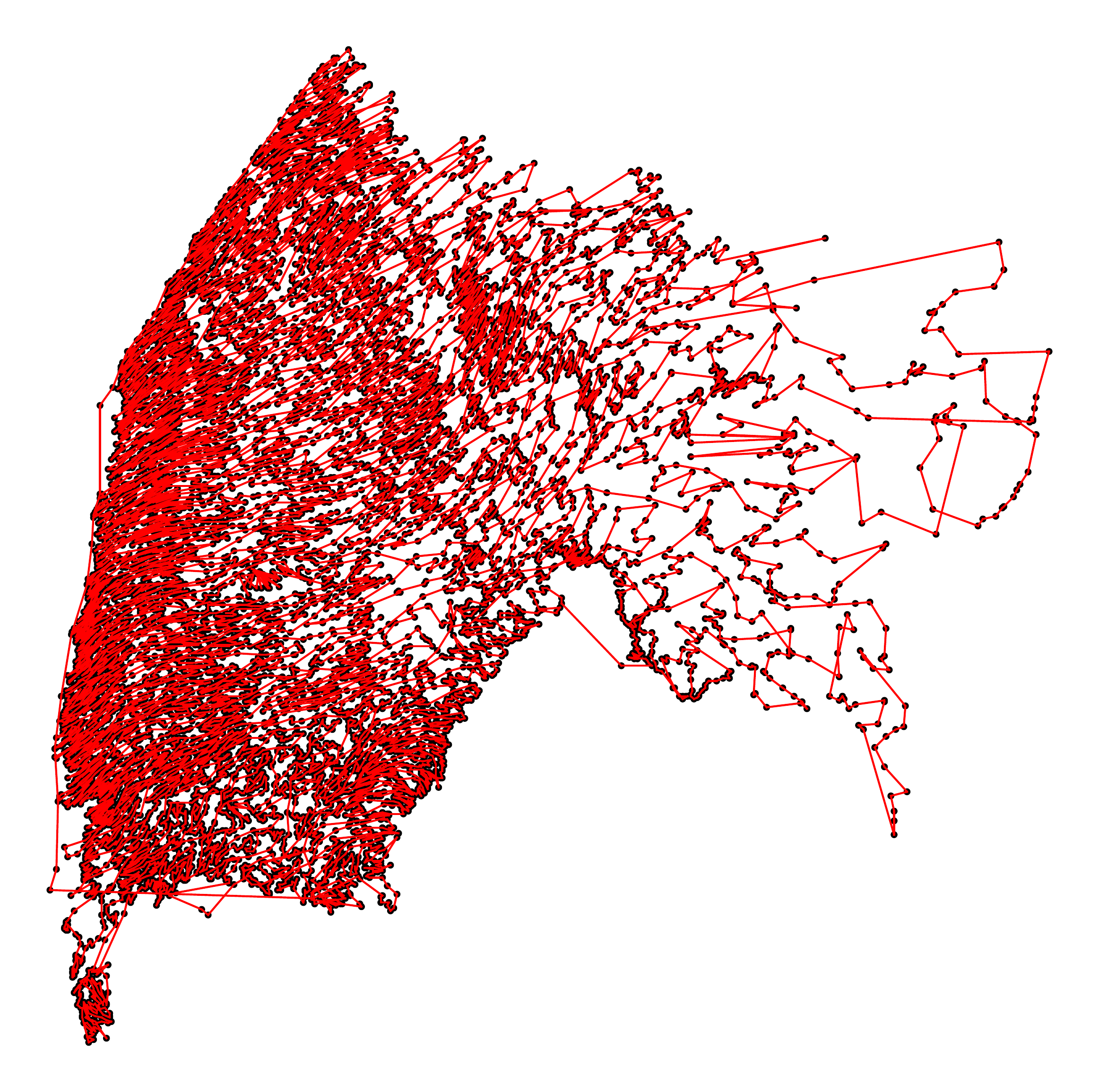}}
\subfigure[INViT: gap=10.04\%]{\includegraphics[width=.8\columnwidth]{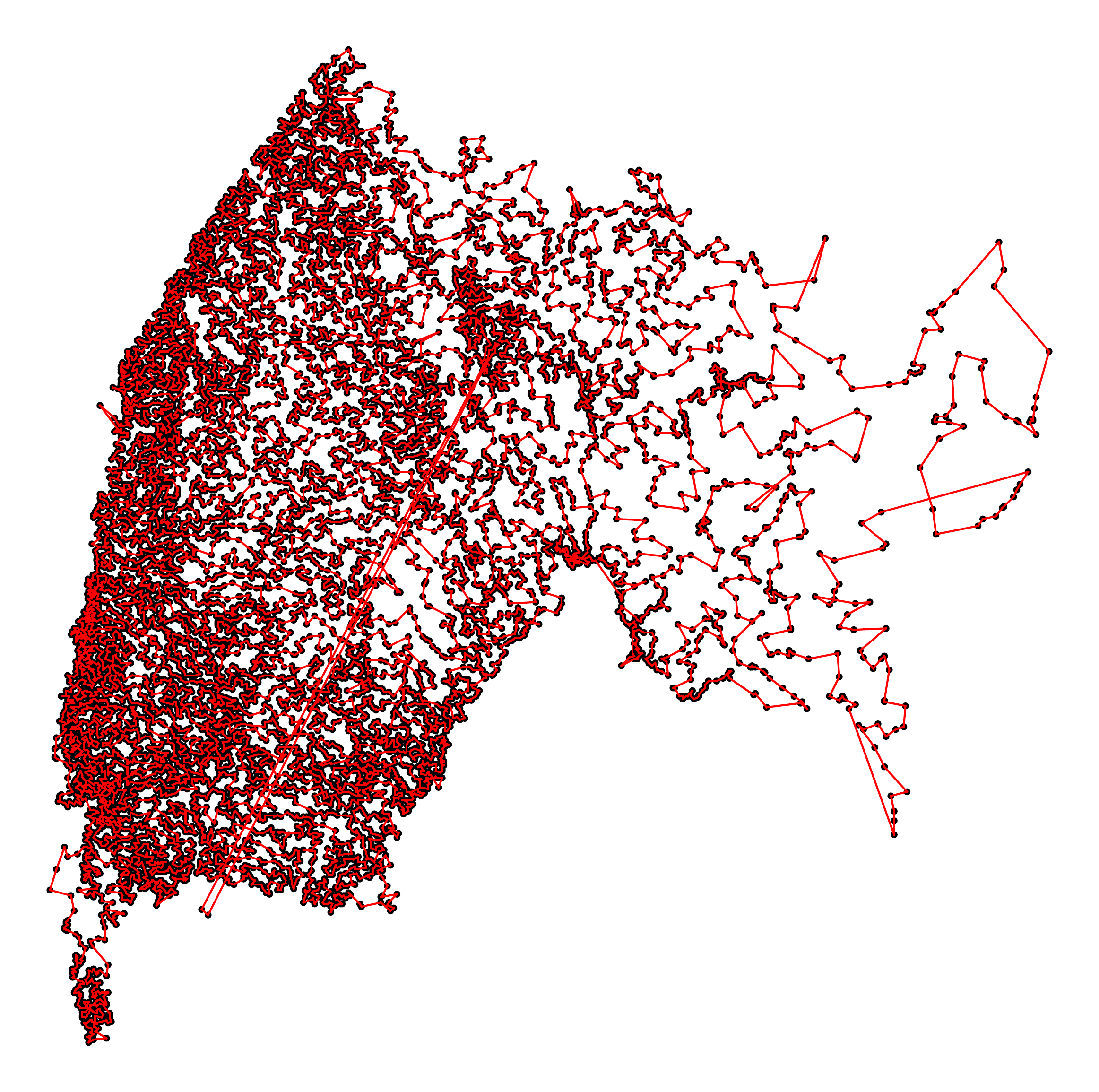}}
\subfigure[GELD (Ours): gap=6.53\%]{\includegraphics[width=.8\columnwidth]{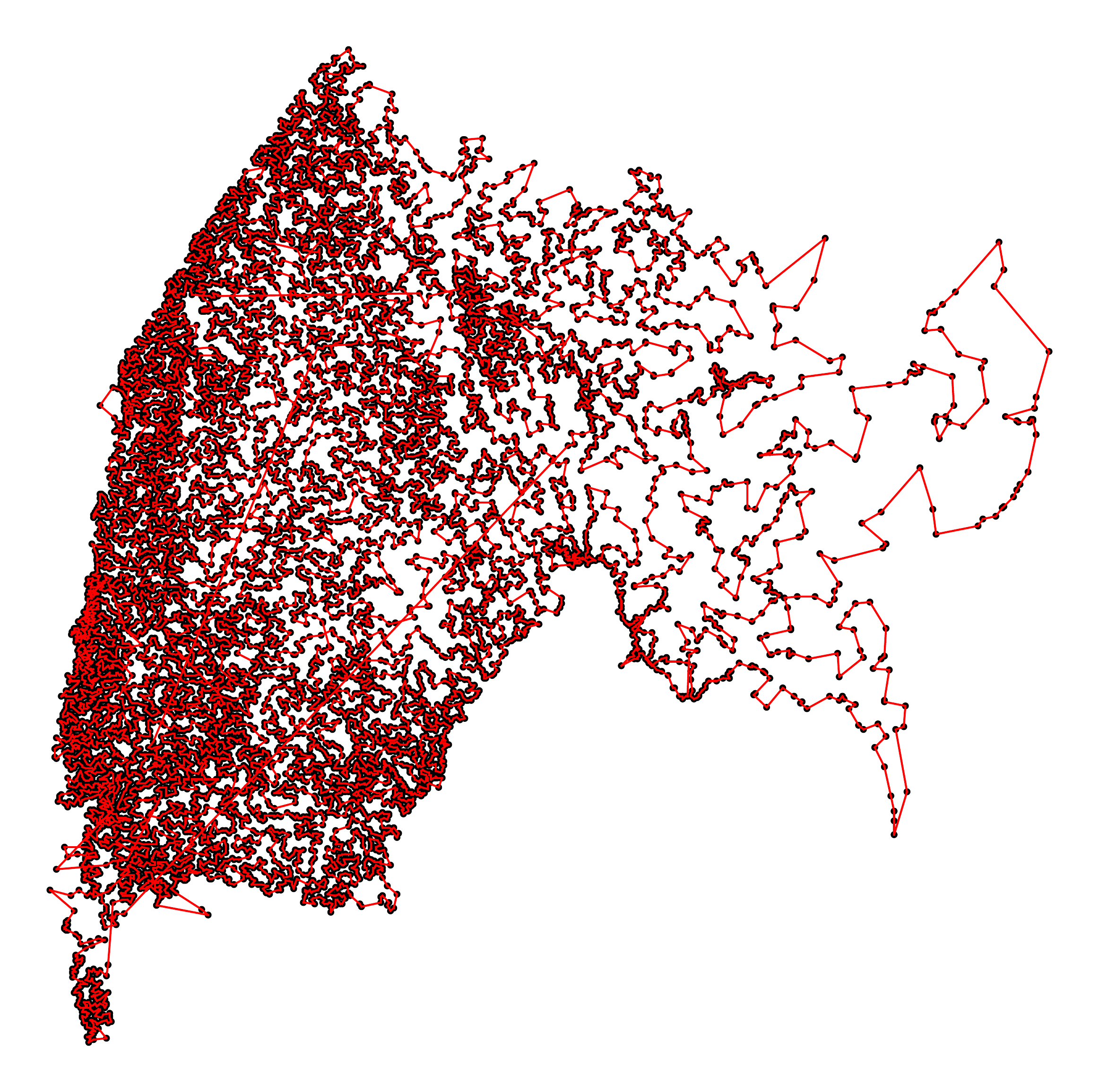}}
\subfigure[RI: gap=13.63\%]{\includegraphics[width=.8\columnwidth]{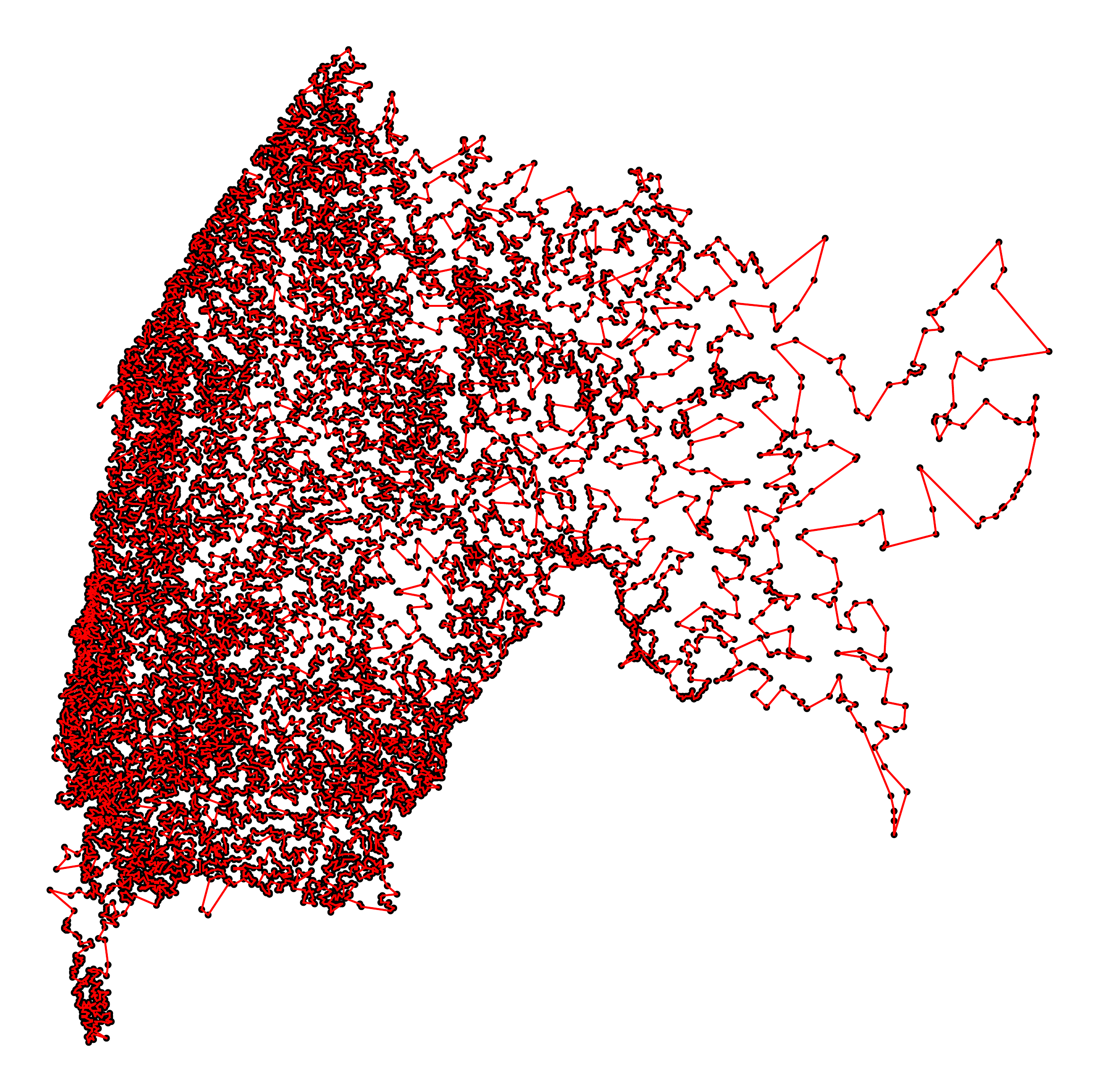}}
\subfigure[RI + GELD (Ours): gap=3.00\%]{\includegraphics[width=.8\columnwidth]{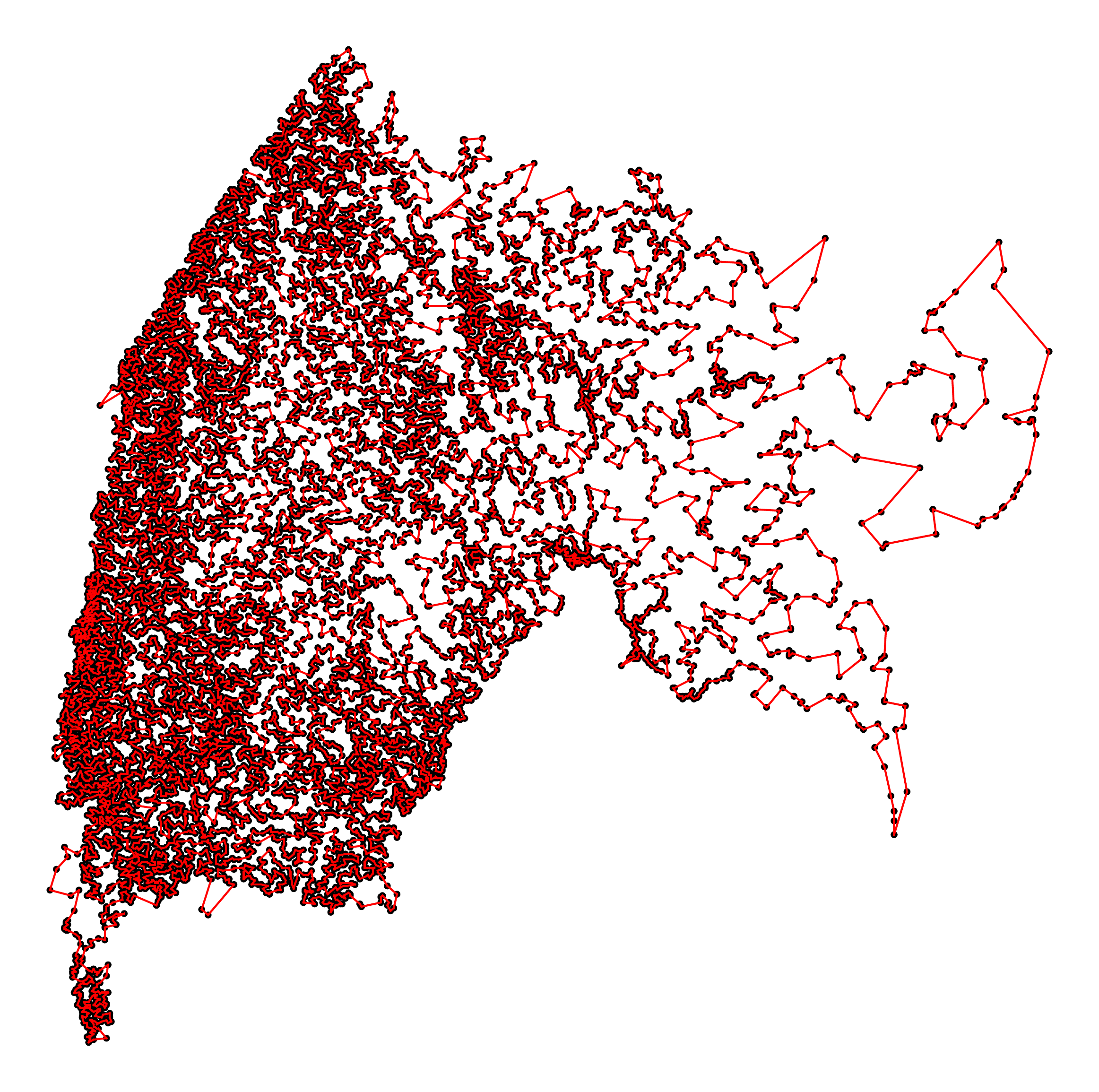}}
\caption{Visualization of solutions on FI10639 TSP instance with 10639 nodes.}\label{fig_FI}
\end{figure*}

Additionally, we visualize the solutions for these two real-world instances in Figures~\ref{fig_TZ} and \ref{fig_FI}, respectively. In each figure, panel (a) displays the optimal solution (i.e., the best-known solution), while panels (b), (c), (d), (e), and (f) show the solutions produced by LEHD (G), INViT-3V (G$^\dagger$), GELD (BOTH), RI, and RI + GELD (PRC(1000)), respectively.

\section{Conclusion}
In this study, we positively answer the proposed research question with ample experimental results as supporting evidence. Specifically, we introduce GELD, which effectively solves TSPs across different scales while capable of exchanging affordable computing time for significantly improved solution quality. We believe the proposed \textit{broad global assessment and refined local selection} framework will offer valuable insights towards solving other COPs. Going forward, we plan to extend the capability of GELD to solve more complex COPs, such as the capacitated vehicle routing problem.











\printcredits

\bibliographystyle{cas-model2-names}

\bibliography{MyBib1}

\appendix
\section{Detailed Results on Real-world Datasets} \label{real_res}
We conduct a comprehensive evaluation of the baseline models and GELD on both TSPLIB and World TSP instances, as detailed in Tables~\ref{det_LIB}-\ref{det_WOR}. Additionally, the performance of baseline models, when integrated with GELD on the World TSP dataset, is presented in Table~\ref{det_WOR_post}.

The largest TSP instance each baseline model can solve is as follows: Omni-TSP (10,639), LEHD (14,051), BQ (11,849), ELG (10,639), INViT-3V (33,708), GD (18,512), and UDC (10,639). Additionally, UDC failed to solve instances with fewer than 100 nodes due to unknown errors.

The results on real-world datasets and synthetic datasets demonstrate GELD outperforms all baseline models, including the SOTA D\&C-based model UDC \citep{Zheng2024}. This superior performance can be attributed to GELD's effective integration of global and local information, whereas UDC is suboptimal in these experiments because it may overlook correlations between sub-problems.

\begin{table*}[!t]
\centering
\caption{Detailed results (gap (\%)) for all included TSPLIB instances}\label{det_LIB}
\begin{tabular}{cccccccccc}
\toprule
 \multirow{2}{*}{Instance} & \multirow{2}{*}{UDC} & \multirow{2}{*}{GD}  & \multirow{2}{*}{INViT-3V} & \multirow{2}{*}{BQ} & \multirow{2}{*}{LEHD} & \multirow{2}{*}{ELG} & \multirow{2}{*}{Omni-TSP} & \multicolumn{2}{c}{GELD (Ours)}   \\
  &  & &  &  &  & &  & G & S$^*$ \\ 
  \hline
  eil51  & -   & 6.66   & 0.94                         & 2.71                        & 1.64                      & 1.41                     & 2.82                          & 1.39        & \textbf{0.70}  \\
berlin52 & - & 0.99   & 0.11                         & 17.08                       & 0.03                      & \textbf{0.01}                     & 12.97                         & 0.04        & 0.03  \\
st70 & - & 0.33       & 1.19                         & 2.06                        & 0.33                      & \textbf{0.15}                     & 2.22                          & 1.63        & 0.31  \\
pr76 & - & 0.99       & 0.36                         & 0.11                        & 0.22                      & 0.69                     & 2.45                          & 0.13        & \textbf{0.00}  \\
eil76 & -  & 2.81     & 2.79                         & 4.92                        & 2.54                      & 1.49                     & 5.20                          & 2.65        & \textbf{1.37}  \\
rat99  & - &0.91     & 1.57                         & 18.49                       & 1.10                      & 4.54                     & 13.13                         & 0.96        & \textbf{0.68}  \\
kroA100 & \textbf{0.02} &0.13    & 0.42                         & 12.15                       & 0.12                      & 1.67                     & 9.07                          & 0.43        & \textbf{0.02}  \\
kroE100 & 0.50  &0.07    & 1.15                         & 13.63                       & 0.43                      & 2.21                     & 5.12                          & 0.57        & \textbf{0.00}  \\
kroB100 & 0.18 &0.45     & 0.26                         & 4.35                        & 0.26                      & 1.65                     & 12.78                         & 0.31        & \textbf{0.00} \\
rd100 & 0.37  &0.15      & 2.48                         & 9.50                        & 0.01                      & 0.44                     & 1.29                          & 1.10        & \textbf{0.01}  \\
kroD100  & 0.07 & 7.24    & 2.18                         & 11.13                       & 0.38                      & 2.62                     & 5.35                          & 1.43        & \textbf{0.00}  \\
kroC100 & \textbf{0.01} &0.64    & 0.34                         & 7.50                        & 0.32                      & 1.87                     & 10.07                         & \textbf{0.01}        & \textbf{0.01}  \\
eil101  & 2.81  & 3.57   & 3.82                         & 4.77                        & 2.31                      & \textbf{0.64}                     & 3.82                          & 2.38        & 2.07  \\
lin105 & \textbf{0.03}  &0.19     & 1.72                         & 12.35                       & 0.34                      & 2.57                     & 11.01                         & 0.19        & \textbf{0.03}  \\
pr107 & 0.65  &4.89      & 1.22                         & 13.74                       & 11.24                     & 3.60                     & 3.66                          & 4.39        & \textbf{0.00}  \\
pr124 & 0.88  &1.78     & 0.53                         & 16.84                       & 1.11                      & 0.26                     & 1.46                          & 21.03       & \textbf{0.08}  \\
bier127 & 1.09 &2.04     & 2.79                         & 6.30                        & 4.76                      & 4.70                     & 8.34                          & 7.55        & \textbf{0.01}  \\
ch130  & \textbf{0.15} &1.11     & 1.90                         & 0.20                        & 0.55                      & 0.43                     & 4.19                          & 1.30        & 0.58  \\
pr136 & 0.42 & \textbf{0.24}      & 1.97                         & 9.87                        & 0.45                     & 2.28                     & 1.04                          & 2.42        & 1.74  \\
pr144 & 0.50  &0.38     & 1.30                         & 14.73                       & \textbf{0.19}                      & 0.55                     & 4.21                          & 2.42        & 0.38  \\
kroA150 & \textbf{0.00}  &0.93    & 1.08                         & 4.95                        & 1.40                      & 2.04                     & 4.91                          & 1.03        & 0.37  \\
kroB150 & 0.08 &0.51    & 2.74                         & 7.19                        & 0.76                      & 1.47                     & 6.02                          & \textbf{0.04}        & \textbf{0.04}  \\
ch150 & 0.37  &0.70     & 2.10                         & 5.64                        & 0.52                      & 1.10                     & 2.45                          & 0.89        & \textbf{0.04}  \\
pr152 & 1.57  &11.53      & 6.63                         & 11.92                       & 12.14                     & \textbf{0.41}                     & 1.20                          & 9.34        & 6.48  \\
u159 & 0.88 &0.92        & 1.84                         & \textbf{0.00}                        & 1.13                      & 1.39                     & 2.06                          & 0.88        & 0.74  \\
rat195 & 0.92 &2.25     & 2.80                         & 10.93                       & 1.42                      & 6.11                     & 19.80                         & 1.50        & \textbf{0.82}  \\
d198  & \textbf{4.44}  &10.34     & 10.44                        & 10.31                       & 9.23                      & 14.23                    & 14.25                         & 13.25       & 6.46  \\
kroA200  & \textbf{0.06} &1.13     & 1.49                         & 8.79                        & 0.64                      & 2.09                     & 6.46                          & 0.84        & 0.16  \\
kroB200  & 0.20 &0.39    & 2.86                         & 10.74                       & \textbf{0.16}                      & 1.58                     & 9.25                          & \textbf{0.16}        & \textbf{0.16}  \\
tsp225  & \textbf{0.00} &0.46      & 1.53                         & 4.70                        & \textbf{0.00}                     & 4.52                     & 8.48                          & 0.16        & 0.00 \\
ts225   & 0.19 &0.33     & 4.68                         & 13.48                       & 0.28                      & 2.52                     & 2.56                          & 1.10        & \textbf{0.00}  \\
pr226  & 0.30  &0.62     & 3.73                         & 11.75                       & 1.11                      & 1.43                     & 2.01                          & 10.72       & \textbf{0.01}  \\
gil262  & 3.38  &\textbf{0.85}     & 2.99                         & 4.76                        & 1.60                      & 2.06                     & 43.99                         & 5.92        & 1.05  \\
pr264   & \textbf{0.15} &16.89     & 3.47                         & 12.50                       & 5.48                      & 5.66                     & 6.17                          & 17.40       & 9.48  \\
a280  & 2.95  &2.34      & 3.88                         & \textbf{0.46}                        & 3.02                      & 5.93                     & 8.72                          & 2.03        & 1.02  \\
pr299  & 2.34  &1.59     & 4.31                         & 6.65                        & 2.81                      & 4.92                     & 10.65                         & 0.69        & \textbf{0.21}  \\
lin318   & 7.10 &1.98    & 3.16                         & 10.36                       & 1.41                      & 4.42                     & 8.17                          & 1.53        & \textbf{0.97}  \\
rd400  & 1.79  &2.36     & 3.91                         & 3.05                        & 1.00                      & 6.26                     & 5.14                          & 3.10        & \textbf{0.52}  \\
f1417  & 7.24  &33.66     & \textbf{4.99}                         & 19.01                       & 7.76                      & 7.55                     & 15.15                         & 20.75       & 7.77  \\
pr439   & 12.87 &3.03     & 7.02                         & 7.14                        & 3.37                      & 7.45                     & 12.06                         & 7.93        & \textbf{1.55}  \\
pcb442  & 4.88 &9.26      & 2.96                         & 0.90                        & 3.11                      & 7.05                     & 8.59                          & 0.35        & \textbf{0.33}  \\
d493  & 7.95   &12.19      & 7.68                         & 8.00                        & 9.49                      & 31.18                    & 27.95                         & 6.20        & \textbf{3.91}  \\
u574  & 4.15  &3.02      & 5.22                         & 1.76                        & 2.73                      & 10.40                    & 18.73                         & 1.37        & \textbf{0.40}  \\
rat575  & 7.78  &8.98     & 4.36                         & 10.07                       & 3.02                      & 9.49                     & 21.48                         & 2.24        & \textbf{0.77}  \\
p654   & 33.07   &22.37     & 10.78                        & 16.03                       & \textbf{3.30}                      & 4.32                     & 14.60                         & 10.04       & 6.41  \\
d657   & 10.25  &4.81     & 8.91                         & 8.62                        & 8.05                      & 11.36                    & 15.09                         & 9.02        & \textbf{1.77}  \\
u724  & 3.75 &4.88       & 3.86                         & 2.18                        & 3.27                      & 10.35                    & 19.35                         & 1.96        & \textbf{0.86}  \\
rat783   & 4.49 &7.11    & 4.85                         & 9.81                        & 3.91                      & 9.56                     & 28.26                         & 1.95        & \textbf{1.28}  \\
pr1002  & \textbf{1.84}  &7.84    & 7.53                         & 8.75                        & 4.44                      & 11.54                    & 20.55                         & 5.85        & 2.80  \\
u1060 & 9.23  &18.00     & 6.39                         & 8.63                        & 10.00                     & 12.18                    & 31.32                         & 12.33       & \textbf{2.87}  \\
vm1084 & 3.75 & 22.47    & 6.24                         & 10.39                       & 5.42                      & 15.81                    & 25.62                         & 3.47        & \textbf{1.18}  \\
pcb1173  & 9.15 & 11.62  & 5.51                         & 11.70                       & 8.01                      & 13.95                    & 27.28                         & 2.38        & \textbf{1.34}  \\

\bottomrule
\end{tabular}
\end{table*}

\begin{table*}[!t]
\centering
\caption{Continued from previous page}
\begin{tabular}{cccccccccc}
\toprule
 \multirow{2}{*}{Instance} & \multirow{2}{*}{UDC} & \multirow{2}{*}{GD}  & \multirow{2}{*}{INViT-3V} & \multirow{2}{*}{BQ} & \multirow{2}{*}{LEHD} & \multirow{2}{*}{ELG} & \multirow{2}{*}{Omni-TSP} & \multicolumn{2}{c}{GELD (Ours)}   \\
  &  & &  &  &  & &  & G & S$^*$ \\ 
  \hline
  d1291  & 12.90  &22.51    & 13.16                        & 11.13                       & 14.13                     & 9.39                     & 32.43                         & 12.44       & \textbf{4.62}  \\
r11304 & 13.59 &15.40     & 6.83                         & 8.77                        & 8.14                      & 13.30                    & 25.62                         & 4.37        & \textbf{1.41}  \\
r11323 & 9.73  & 18.19    & 6.75                         & 7.64                        & 9.26                      & 12.42                    & 29.76                         & 12.59       & \textbf{2.27}  \\
nrw1379  & 9.57 &104.77   & 4.38                         & 9.83                        & 15.49                     & 12.57                    & 23.00                         & 2.27        & \textbf{1.00}  \\
f1400 & 25.11   &84.65    & 11.89                        & 31.19                       & 18.80                     & 8.74                     & 18.18                         & 23.12       & \textbf{7.15}  \\
u1432 & 6.61 &10.30      & 4.25                         & 4.98                        & 7.96                      & 10.65                    & 22.30                         & 5.07        & \textbf{2.80}  \\
f1577  & 23.75  &65.74     & 7.53                         & 21.61                       & 14.68                     & 8.35                     & 32.75                         & 9.44        & \textbf{5.15}  \\
d1655& 9.11   &47.28     & 10.58                        & 17.01                       & 13.89                     & 15.66                    & 34.92                         & 14.10       & \textbf{6.45}  \\
vm1748 & 7.68  &19.12    & 8.41                         & 11.18                       & 10.10                     & 17.13                    & 30.84                         & 4.35        & \textbf{0.86}  \\
u1817 & 8.39  &28.70     & 6.90                         & 9.43                        & 10.32                     & 12.62                    & 39.72                         & 9.43        & \textbf{3.08}  \\
rl1889 & 22.28 &26.59     & 9.08                         & 14.91                       & 7.49                      & 17.12                    & 37.50                         & 6.32        & \textbf{3.41}  \\
d2103 & 17.96  &57.66     & 10.48                        & 17.47                       & 14.57                     & 6.90                     & 36.05                         & 10.88       & \textbf{4.42}  \\
u2152 & 13.55   &32.67     & 7.20                         & 9.08                        & 12.65                     & 12.12                    & 43.01                         & 8.68        & \textbf{5.16}  \\
u2319 & 6.06   &19.98    & 0.62                         & 3.41                        & 4.18                      & 3.88                     & 17.61                         & 0.43        & \textbf{0.34}  \\
pr2392 & 11.17  &32.68    & 6.80                         & 9.26                        & 12.33                     & 16.95                    & 40.08                         & 6.12        & \textbf{3.04}  \\
pcb3038 & 7.14  &35.92    & 7.05                         & 13.44                       & 13.44                     & 16.75                    & 40.08                         & 8.63        & \textbf{2.73}  \\
fl3795  & 40.23  & 331.22    & 11.29                        & 32.09                       & 13.55                     & 13.46                    & 54.24                         & 21.26       & \textbf{10.66} \\
fn14461 & 17.29 &134.34    & 5.58                         & 21.38                       & 19.05                     & 15.98                    & 47.99                         & 12.38       & \textbf{2.99}  \\
r15915 & 21.10  & 288.03   & 8.68                         & 24.58                       & 24.17                     & 16.17                    & 62.61                         & 11.83       & \textbf{7.02}  \\
r15934 & 31.41 &363.20     & 10.00                        & 30.17                       & 24.11                     & 18.08                    & 63.94                         & 11.68       & \textbf{7.17}  \\
r111849  & 23.37 &598.01     & 9.05                         & 45.21                       & 38.04                     & OOM                        & OOM                             & 14.94       & \textbf{6.11}  \\
usa13509 & OOM  &2252.54   & \textbf{8.23}                         & OOM                           & 71.11                     & OOM                        & OOM                             & 17.39       & 8.97  \\
brd14051 & OOM &700.75    & 7.40                         & OOM                           & 41.22                     & OOM                        & OOM                             & 17.32       & \textbf{4.17}  \\
d15112 & OOM &660.57      & 6.21                         & OOM                           & OOM                         & OOM                        & OOM                             & 14.57       & \textbf{3.58}  \\
d18512 & OOM  & 744.35     & 6.99                         & OOM                           & OOM                         & OOM                        & OOM                             & 15.26       & \textbf{6.64}  \\
\hline
Avg. gap & 7.34 & 90.34  & 4.86                         & 10.92                       & 7.56                      & 7.25                     & 18.07                         & 6.28        & \textbf{2.35}  \\
Avg. time  & 5.6s & 2.7m & 26.2s                          & 1.4m                          & 47.0s                       & 6.1s                       & 3.8s                            & 3.8s          & 27.6s\\
\bottomrule
\end{tabular}
\end{table*}

\begin{table*}[!t]
\centering
\caption{Detailed results (gap(\%)) for all included National TSPs} \label{det_WOR}
\begin{tabular}{cccccccccc}
\toprule
 \multirow{2}{*}{Instance} & \multirow{2}{*}{UDC} & \multirow{2}{*}{GD}  & \multirow{2}{*}{INViT-3V} & \multirow{2}{*}{BQ} & \multirow{2}{*}{LEHD} & \multirow{2}{*}{ELG} & \multirow{2}{*}{Omni-TSP} & \multicolumn{2}{c}{GELD (Ours)}   \\
  &  & &  &  &  & &  & G & S$^*$ \\ 
  \hline
WI29 & -  & 0.60      & 0.00   & 19.95  & 0.06   & 4.54   & 0.05   & 0.71        & \textbf{0.00} \\
DJ38 & -   &6.41     & 0.06   & 28.63  & 0.17   & \textbf{0.02}   & 5.21   & 0.11        & 0.05  \\
QA194 & 0.58 &236.40     & 2.88   & 12.18  & 27.15  & 7.06   & 10.44  & 0.53        & \textbf{0.02}  \\
UY734 & 5.54  &284.43      & 5.38   & 9.26   & 20.98  & 10.77  & 14.83  & 3.35        & \textbf{2.00}  \\
ZI929 & 12.80 &111.47      & 6.54   & 13.67  & 18.34  & 14.61  & 21.24  & 9.05        & \textbf{2.97}  \\
LU980 & 11.78  &2368.93     & 4.96   & 7.83   & 93.28  & 12.27  & 17.58  & 2.89        & \textbf{1.40}  \\
RW1621 & 17.34 &2722.21     & 7.42   & 12.79  & 58.53  & 11.42  & 27.20  & 7.99        & \textbf{3.61}  \\
MU1979 & 15.41   & 1351.43   & 13.06  & 48.92  & 42.65  & 22.54  & 52.06  & 17.55       & \textbf{9.06}  \\
NU3496 & 17.57 &3616.20     & 10.74  & 22.12  & 84.94  & 17.20  & 44.75  & 10.91       & \textbf{3.58}  \\
CA4663 & 22.39 & 684.98    & \textbf{9.47}   & 78.38  & 40.22  & 88.57  & 162.60 & 22.77       & 11.64 \\
TZ6117 & 23.26  &2007.00    & 9.45   & 32.11  & 51.24  & 20.69  & 59.20  & 14.68       & \textbf{5.64}  \\
EG7146 & 27.11  &1281.81    & 12.88  & 170.87 & 42.15  & 209.58 & 151.05 & 19.06       & \textbf{5.55}  \\
YM7663 & 28.03  &3632.62     & 13.37  & 82.37  & 93.84  & 60.12  & 79.25  & 18.06       & \textbf{7.17}  \\
PM8079 & 20.85  & 7728.54    & 10.47  & 103.36 & 207.10 & 22.85  & 72.56  & 17.00       & \textbf{7.52}  \\
EI8246  & 15.70  & 4568.96   & \textbf{7.09}   & 39.19  & 131.26 & 20.73  & 61.70  & 14.98       & 7.73  \\
AR9152 & 29.17  &1753.86     & 12.63  & 64.43  & 56.54  & 21.66  & 72.70  & 16.74       & \textbf{9.10}  \\
JA9847 & 46.86  &10429.63    & \textbf{12.20}  & 197.73 & 132.74 & 37.71  & 80.34  & 23.58       & 12.61 \\
GR9882 & 19.44 &2245.37     & 13.25  & 78.22  & 74.69  & 23.95  & 70.10  & 18.57       & \textbf{6.10}  \\
KZ9976 & 18.58  &1172.08     & 9.15   & 86.37  & 55.72  & 23.30  & 102.23 & 19.25       & \textbf{7.55}  \\
FI10639 & 18.14  &3709.73   & 10.04  & 55.65  & 98.52  & 22.44  & 71.67  & 14.62       & \textbf{6.53}  \\
MO14185 & OOM &3629.80    & 8.41   & OOM      & OOM      & OOM      & OOM      & 16.07       & \textbf{7.07}  \\
HO14473 & OOM &9842.77    & 13.13  & OOM      & OOM      & OOM      & OOM      & 18.35       & \textbf{8.16}  \\
IT16862 & OOM &3762.62     & 9.13   & OOM      & OOM      & OOM      & OOM      & 18.88       & \textbf{7.39}  \\
VM22775& OOM   & OOM    & 9.75   & OOM      & OOM      & OOM      & OOM      & 20.41       & \textbf{8.32}  \\
SW24978 & OOM  & OOM  & 8.58   & OOM      & OOM      & OOM      & OOM      & 17.87       & \textbf{6.88}  \\
BM33708 & OOM  & OOM    & \textbf{7.35}   & OOM      & OOM      & OOM      & OOM      & 18.76       & 7.42  \\
CH71009 & OOM  & OOM   & OOM      & OOM      & OOM      & OOM      & OOM      & 25.41       & \textbf{13.92} \\
\hline
Avg. gap & 19.44 & 2919.47  & 8.75   & 58.20  & 66.51  & 32.60  & 58.83  & 14.39       & \textbf{6.26}  \\
Avg. time  & 6.3s & 8.9m & 3.4m     & 14.8m     & 2.8m     & 3.7m     & 2.3m     & 23.4s         & 1.4m  \\
\bottomrule
\end{tabular}    
\end{table*}

\begin{table*}[h]
\centering
\caption{Detailed results (gap(\%)) for all included National TSPs using GELD (with PRC(1000)) as a post-processing method}\label{det_WOR_post}
\begin{tabular}{cccccccccc}
\toprule
 \multirow{2}{*}{Instance} &UDC  &GD   & INViT-3V & BQ & LEHD & ELG & Omni-TSP & \multicolumn{2}{c}{RI}   \\
 & +Ours & +Ours & +Ours  & +Ours & +Ours &+Ours& +Ours & - & +Ours \\ 
 \hline
WI29 & - &0.53       & 0.00 & 0.00 & 0.00 & 4.25  & 0.00 & 0.00 & 0.00 \\
DJ38 & -   & 0.05     & 0.05  & 0.05  & 0.05  & 0.05  & 0.05  & 17.55 & 0.05  \\
QA194 & 0.58   & 4.52    & 0.68  & 4.52  & 7.12  & 0.67  & 0.67  & 11.54 & 2.35  \\
UY734 & 4.07   &  20.72   & 1.37  & 2.88  & 2.90  & 1.91  & 1.51  & 13.23 & 1.26  \\
ZI929 & 11.87  & 13.43    & 2.95  & 5.91  & 3.61  & 4.30  & 1.92  & 9.35  & 2.68  \\
LU980 & 6.83  & 33.83     & 2.42  & 1.64  & 3.21  & 1.48  & 1.39  & 12.00 & 1.30  \\
RW1621 & 12.71   & 70.19    & 1.98  & 1.26  & 4.84  & 1.74  & 3.55  & 12.48 & 1.41  \\
MU1979 & 11.33   & 70.43   & 7.22  & 14.58 & 8.65  & 7.27  & 5.34  & 9.09  & 2.39  \\
NU3496  & 7.82  & 94.52   & 3.65  & 4.77  & 10.14 & 4.39  & 4.53  & 13.58 & 3.83  \\
CA4663  & 13.25 & 27.96    & 5.43  & 7.42  & 9.97  & 21.71 & 8.54  & 14.81 & 4.94  \\
TZ6117  & 13.44 & 87.03    & 3.89  & 8.73  & 8.15  & 6.12  & 4.64  & 14.42 & 2.76  \\
EG7146 & 20.14   & 60.27   & 6.56  & 7.93  & 7.58  & 63.24 & 18.72 & 14.35 & 4.07  \\
YM7663  & 17.43 & 207.35    & 8.59  & 10.19 & 10.82 & 25.99 & 7.24  & 13.79 & 3.68  \\
PM8079 & 9.79 & 203.51     & 2.53  & 7.15  & 11.90 & 8.41  & 5.83  & 12.07 & 3.12  \\
EI8246 & 9.95 &  188.51    & 3.02  & 7.71  & 10.70 & 7.92  & 4.24  & 14.14 & 3.31  \\
AR9152 & 18.64    & 141.87   & 7.46  & 9.18  & 8.39  & 9.16  & 6.23  & 13.73 & 3.58  \\
JA9847 & 34.18  & 135.25    & 5.58  & 15.65 & 7.08  & 16.33 & 5.94  & 12.80 & 3.89  \\
GR9882 & 13.62 & 97.99     & 7.13  & 9.78  & 9.43  & 8.22  & 3.03  & 12.02 & 1.92  \\
KZ9976 & 10.89 & 58.46     & 3.36  & 9.94  & 9.68  & 7.62  & 8.30  & 14.02 & 3.32  \\
FI10639 & 11.19 & 158.62    & 5.49  & 8.81  & 11.64 & 7.99  & 5.40  & 13.63 & 3.00  \\
MO14185 & -  & 106.41   & 3.41  & -       & -       & -       & -       & 13.45 & 2.86  \\
HO14473 & - & 194.83   & 7.69  & -       & -       & -       & -       & 11.68 & 3.11  \\
IT16862 & - &103.73    & 4.71  & -       & -       & -       & -       & 13.77 & 2.66  \\
VM22775 & - & -    & 5.50  & -       & -       & -       & -       & 12.44 & 2.19  \\
SW24978 & - & -    & 3.55  & -       & -       & -       & -       & 13.87 & 3.02  \\
BM33708 & - & -    & 2.81  & -       & -       & -       & -       & 13.72 & 2.77  \\
CH71009 & - & -    & -       & -       & -       & -       & -       & 14.21 & 3.61  \\
\hline
Avg. gap & 12.66 & 90.43  & 4.12  & 6.91  & 7.29  & 10.44 & 4.85  & 12.66 & 2.71  \\
Gain(\%) &\textbf{35.05} & \textbf{96.90}  & \textbf{52.91}  & \textbf{88.13}  & \textbf{89.04}  & \textbf{67.98} & \textbf{91.76}  & \multicolumn{2}{c}{\textbf{78.59}}  \\

Avg. time & + 26.2s& +28.3s & +34.0s  & +26.6s  & +26.6s  & +26.6s  & +26.6s  & 1.1s    & +36.8s  \\
\bottomrule
\end{tabular}
    
\end{table*}



\end{document}